\pdfoutput=1

\documentclass[11pt]{article}

\usepackage{acl}

\usepackage{times}
\usepackage{latexsym}
\usepackage{graphicx}
\usepackage{float}

\usepackage[T1]{fontenc}

\usepackage[utf8]{inputenc}

\usepackage{microtype}
\usepackage{booktabs}
\usepackage{multicol}
\usepackage{multirow}
\usepackage{tcolorbox}
\usepackage{enumitem}
\usepackage{xcolor}
\usepackage{xspace}
\usepackage{amsmath}
\usepackage{subcaption}
\usepackage{tabularx}
\usepackage{longtable}
\usepackage{array}
\usepackage{float}
\usepackage{placeins} 
\usepackage{xcolor} %
\usepackage{listings} %

\definecolor{pastelgreen}{HTML}{97D077}
\definecolor{customisedblue}{HTML}{6C8EBF}
\definecolor{lightblue}{HTML}{DAE8FC}

\newenvironment{itemizesquish}[2]{\begin{list}{\labelitemi}{\setlength{\itemsep}{#1}\setlength{\labelwidth}{#2}\setlength{\leftmargin}{\labelwidth}\addtolength{\leftmargin}{\labelsep}}}{\end{list}}

\def\ie{{\em i.e.,}\xspace}

\usepackage{soul}

\title{CoMAT: Chain of Mathematically Annotated Thought Improves Mathematical Reasoning}

\author{
 \textbf{Joshua Ong Jun Leang$^{1,2}$} \quad
 \textbf{Aryo Pradipta Gema$^{1}$} \quad
 \textbf{Shay B. Cohen$^{1}$}
\\
$^{1}$ School of Informatics, The University of Edinburgh \quad
$^{2}$Imperial College London 
\medskip
\\
 \texttt{\{jong2,aryo.gema,scohen\}@ed.ac.uk}
}

\begin{document}
\maketitle
\begin{abstract}

Mathematical reasoning remains a significant challenge for large language models (LLMs), despite progress in prompting techniques such as Chain-of-Thought (CoT). We present \textbf{Chain of Mathematically Annotated Thought (CoMAT)}, which enhances reasoning through two stages: \emph{Symbolic Conversion} (converting natural language queries into symbolic form) and \emph{Reasoning Execution} (deriving answers from symbolic representations). CoMAT operates entirely with a single LLM and without external solvers. Across four LLMs, CoMAT outperforms traditional CoT on six out of seven benchmarks, achieving gains of 4.48\% on MMLU-Redux (MATH) and 4.58\% on GaoKao MCQ. In addition to improved performance, CoMAT ensures faithfulness and verifiability, offering a transparent reasoning process for complex mathematical tasks\footnote{Code is available at \url{https://github.com/joshuaongg21/CoMAT/}}.
\end{abstract}

\section{Introduction}

\begin{figure}[t]
    \centering
    \includegraphics[width=0.9\linewidth]{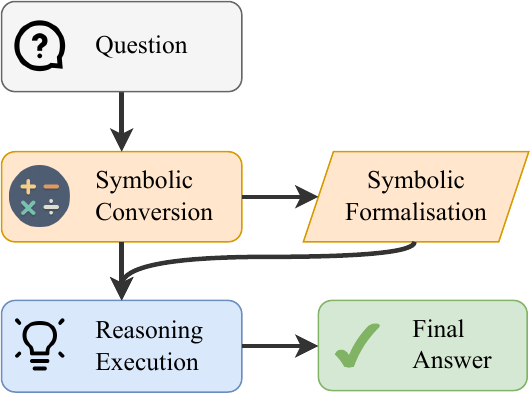}
    \caption{An overview of our CoMAT framework. CoMAT divides complex reasoning tasks into two stages: \emph{Symbolic Conversion}, where queries are translated into structured symbolic reasoning chains (Figure~\ref{fig:CoMAT_main_a}), and \emph{Reasoning Execution}, where step-by-step calculations are performed to derive the final answer (Figure~\ref{fig:CoMAT_main_b}).}
    \label{fig:CoMAT_overview_new}
\end{figure}

Complex mathematical reasoning remains a significant challenge for large language models \cite[LLMs;][]{ luo2024improve,li2023bert,10.1162/tacl_a_00594}.
Techniques like Chain-of-Thought (CoT) prompting~\cite{wei2022chain, kojima2022large, wang2022self} have improved LLM performance by encouraging the generation of intermediate reasoning steps.
However, CoT explanations are not always faithful to the actual reasoning process of the model~\cite{bentham2024chain, turpin2024language, yee2024dissociation}; with final answers that may not logically follow from the reasoning chain, suggesting that LLMs can fabricate reasoning paths~\cite{lyu2023faithful}.

\begin{figure*}[t]
    \centering
    \begin{subfigure}[t]{\linewidth}
        \centering
        \includegraphics[width=\linewidth]{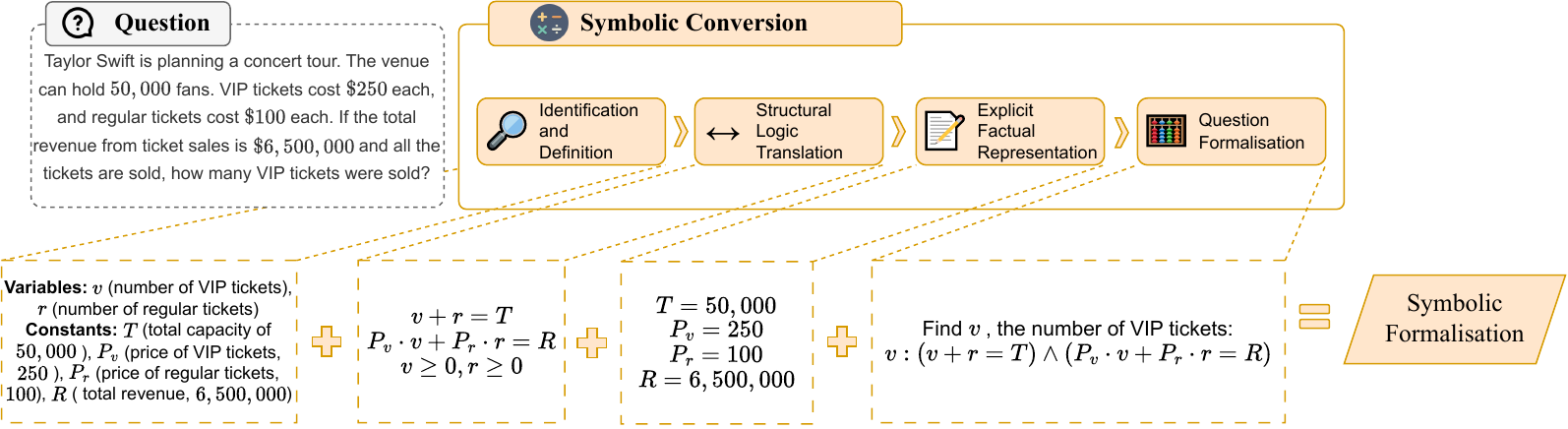}
        \caption{An overview of the Symbolic Conversion stage of CoMAT, which includes \emph{Identification and Definition}, \emph{Structural Logic Translation}, \emph{Explicit Factual Representation}, and \emph{Question Formalisation}.}
        \label{fig:CoMAT_main_a}
    \end{subfigure}
    \vspace{0.1cm}
    \begin{subfigure}[t]{\linewidth}
        \vspace{0.3cm}
        \centering
        \includegraphics[width=\linewidth]{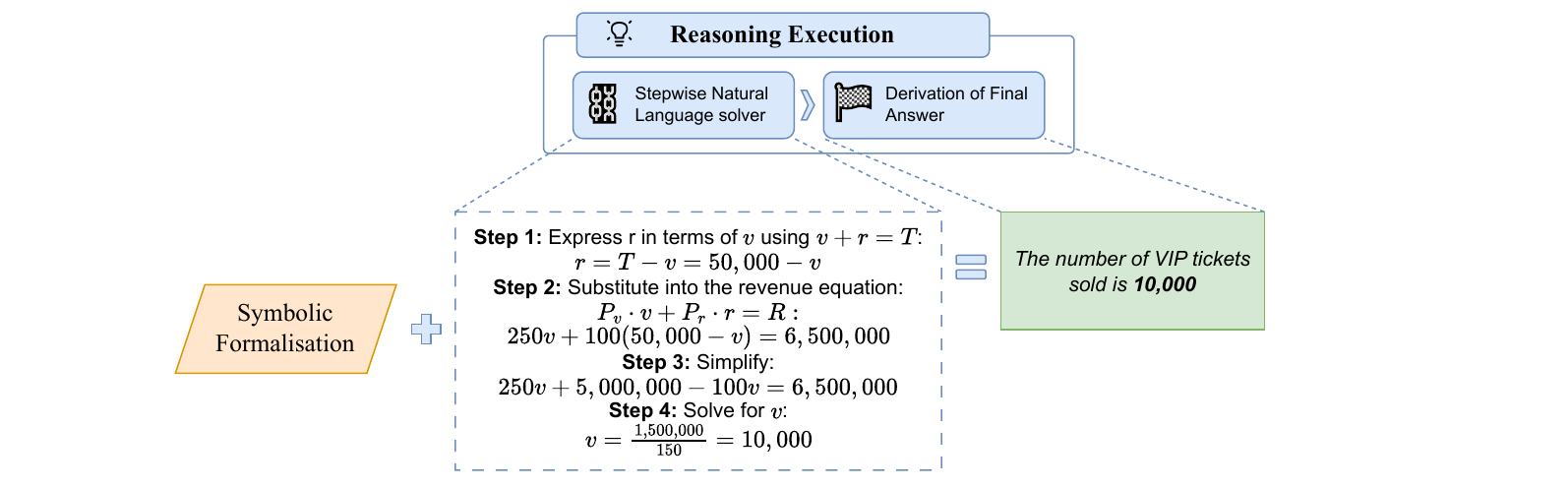}
        \caption{An overview of the Reasoning Execution stage of CoMAT, which performs `step by step' reasoning based on the symbolic representation to provide the final answer.}
        \label{fig:CoMAT_main_b}
    \end{subfigure}
    \caption{An overview of CoMAT divided into two main stages: Symbolic Conversion and Reasoning Execution 
    }
    \label{fig:CoMAT_main_diagram}
\end{figure*}

Recent efforts to improve reasoning faithfulness have relied on integrating external solvers~\cite{lyu2023faithful, he2023solving,jiang2024leanreasoner,leang-etal-2025-theorem}. However, these approaches introduce new challenges: LLMs often generate syntactically invalid code when translating natural language into formal statements, causing solvers to fail~\cite{olausson2023linc, gou2023critic, wen2024fixing, quan2024verification, leang-etal-2025-theorem}. While existing methods incorporate symbolic reasoning without relying on external solvers, they primarily focus on other domains, such as common mathematical benchmarks~\cite{wang2023exploring}, logical reasoning~\cite{jiang2024leanreasoner, xu2024faithful}, and other areas~\cite{li2023symbolic}. As a result, their applicability to challenging and multilingual mathematical reasoning benchmarks remains limited and impractical (see \S\ref{Faithful CoT vs CoMAT}). 

To address these issues, we propose \textbf{Chain of Mathematically Annotated Thought (CoMAT)}, a novel approach that leverages symbolic reasoning entirely within LLMs to tackle a wide range of mathematical reasoning tasks. By eliminating external solvers, CoMAT avoids issues related to code generation failures, offering a more robust solution for a broad range of mathematical tasks. Notably, CoMAT performs well where there are difficulties with code-based methods, such as multilingual datasets and Olympiad-level problems. As shown in Appendix~\ref{Faithful CoT vs CoMAT}, code-based methods often struggle with Olympiad-level mathematics, while CoMAT demonstrates greater adaptability.
\emph{To the best of our knowledge, CoMAT is the first method to apply symbolic reasoning across diverse mathematical reasoning tasks, including multilingual datasets, Olympiad-level problems, and common benchmarks.}

Empirically, CoMAT outperforms most existing baselines. When evaluated on datasets ranging from standard mathematical problems to complex Olympiad-level challenges in both English and Mandarin, CoMAT showed significant improvements. When applied to Gemini-1.5-Pro, we achieved an overall exact match boost of 3.54\% on AQUA and an 8.18\% increase in accuracy on the Olympiad Bench (English) compared to CoT. Similarly, using Qwen2-72B as a baseline, we observed a 2.55\% improvement averaged across mathematical subjects of the MMLU-Redux benchmark. 

To summarise, our key contributions are: %
\begin{itemizesquish}{-0.3em}{0.5em} %
    \item We introduce CoMAT, a novel approach that leverages symbolic reasoning entirely within LLMs, eliminating the need for external tools or verifiers. This ensures both accuracy and verifiability through a transparent and structured reasoning process.
    \item Our symbolic prompts are standardised to tackle a wide array of mathematical reasoning tasks, achieving state-of-the-art performance in selected datasets, as well as competitive results across diverse benchmarks of varying complexity and languages.
    \item We perform a comprehensive ablation study demonstrating the generalisability of CoMAT across multiple LLMs, datasets, prompt designs, and languages.
\end{itemizesquish}

\section{CoMAT: Chain of Mathematically Annotated Thought}
Traditional CoT methods in mathematical reasoning, which rely heavily on natural language explanations, can introduce ambiguity and inconsistencies, leading models to fabricate reasoning paths or misinterpret variable relationships, ultimately compromising accuracy in complex tasks~\cite{lu2024mitigating}.
CoMAT addresses these limitations by adopting a structured, symbolic approach that enforces mathematical consistency and reduces ambiguity.
By using well-defined symbolic logic, CoMAT ensures that each reasoning step adheres to sound mathematical principles, enhancing the model's ability to solve problems accurately.

CoMAT extends traditional CoT by incorporating symbolic transformations as the core component of the reasoning process. While CoT typically follows a $(Q, R, A)$ structure — where $Q$ is the question, $R$ is the reasoning, and $A$ is the final answer — CoMAT introduces a more structured process: $(Q, S, R, A)$, where $S = (s_1, s_2, s_3, s_4)$ represents the four steps in the symbolic reasoning pipeline, designed to break down complex problems into formal, interpretable sequences of logical operations. This structured decomposition enhances transparency and allows for systematic verification of each step. Examples of manual annotation for verifiability can be found in Appendix~\ref{sec:Manual Annotation for CoMAT Verification}. 
\subsection{CoMAT's Two-Stage Process}
\label{CoMAT's Two-Stage Process}
CoMAT consists of two main stages: \textbf{Symbolic Conversion} and \textbf{Reasoning Execution}. Each stage is critical to ensuring the accuracy and faithfulness of the reasoning process.

\paragraph{Symbolic Conversion.}

In this stage, the LLM transforms a natural language query $Q$ into a symbolic representation, $S = (s_1, s_2, s_3, s_4)$. This involves \emph{identification and definition}, \emph{structural logic translation}, \emph{explicit factual representation}, and \emph{question formalisation}, all carried out by the LLM. This stage acts as the foundation for accurate reasoning by converting ambiguous natural language into well-structured mathematical logic.

\paragraph{Reasoning Execution.}
Once the problem is translated into its symbolic form ($S$), the model applies logical reasoning to derive the solution, $S = (s_5, s_6, s_7)$. The logical reasoning, $S_{5}$, is followed by stepwise reasoning, similar to \citet{kojima2022large} --\ie 
``Let's think step-by-step''. 
Incorporating tools in $S_{5}$ may impact mathematical performance due to significant execution errors, which stepwise reasoning in $S_{5}$ effectively mitigates by providing a more efficient and robust framework~\cite{olausson2023linc, gou2023critic}. By grounding the reasoning in the symbolic structure, CoMAT ensures that each step aligns with mathematical logic, reducing the risk of errors or illogical steps. The final answer, $A$, is then generated based on this rigorous reasoning process.

\subsection{Case Study}
To demonstrate how CoMAT works in practice, consider the following math problem related to ticket sales, based on Figure~\ref{fig:CoMAT_main_diagram}:

\begin{tcolorbox}[colback=white,colframe=customisedblue] 
\textit{Taylor Swift is planning a concert tour. The venue can hold 50,000 fans. VIP tickets cost \$250 each, and regular tickets cost \$100 each. If the total revenue from ticket sales is \$6,500,000 and all tickets are sold, how many VIP tickets were sold?} \end{tcolorbox}

\begin{enumerate}[leftmargin=*] %

\item \textbf{Identification and Definition.} CoMAT first identifies and defines the relevant variables and constants to ensure precision in the symbolic reasoning process. For instance:

 \end{enumerate}
 
\begin{tcolorbox}[colback=white,colframe=customisedblue]
\setlist{noitemsep}
\begin{itemize}[leftmargin=*]
\item \textbf{Variables}: $v$ (number of VIP tickets), $r$ (number of regular tickets) 
\item \textbf{Constants}: $T$ (total capacity of 50,000), $P_v$ (price of VIP tickets, $250$), $P_r$ (price of regular tickets, $100$), $R$ (total revenue, $6,500,000$) \end{itemize}
\end{tcolorbox}

\begin{enumerate}[leftmargin=*]

\item[2.] \textbf{Structural Logic Translation}: Next, CoMAT extracts the key variables and translates the problem into formal rules that define their relationships, ensuring the reasoning process is grounded in well-defined constraints:

\end{enumerate}

\begin{tcolorbox}[colback=white,colframe=customisedblue]
    \setlist{noitemsep}
    \begin{itemize}[leftmargin=*]
        \item $v + r = T$
        \item $P_v \cdot v + P_r \cdot r = R$
        \item $v \geq 0, r \geq 0$ (non-negative constraints)
    \end{itemize}
\end{tcolorbox}

\begin{enumerate}[leftmargin=*]

\item[3.] \textbf{Explicit Factual Representation}: CoMAT then integrates all relevant facts into the logical structure to avoid omitting key information:

\end{enumerate}

\begin{tcolorbox}[colback=white,colframe=customisedblue]
    \setlist{noitemsep}
    \begin{itemize}[leftmargin=*]
        \item $T = 50,000$
        \item $P_v = 250$
        \item $P_r = 100$
        \item $R = 6,500,000$
    \end{itemize}
\end{tcolorbox}

\begin{enumerate}[leftmargin=*]

\item[4.] \textbf{Question Formalisation}: CoMAT formalises the question into a symbolic expression to ensure that the reasoning process remains objective and free of bias from answer options. However, the model may inherently choose to parse the question in natural language rather than a symbolic representation. In this case:

\end{enumerate}

\begin{tcolorbox}[colback=white,colframe=customisedblue]
In our example, we are tasked with finding $v$, the number of VIP tickets: 

$\text{Find } v : (v + r = T) \land (P_v \cdot v + P_r \cdot r = R)$
\end{tcolorbox}

\begin{enumerate}[leftmargin=*]

\item[5.] \textbf{Reasoning Execution}: The problem is then solved step-by-step using the symbolic representation, as demonstrated:

\end{enumerate}

\begin{tcolorbox}[colback=white,colframe=customisedblue]
\textbf{Step 1:} Express $r$ in terms of $v$ using $v + r = T$:
        \[
        r = T - v = 50,000 - v
        \]
\textbf{Step 2:} Substitute into the revenue equation $P_v \cdot v + P_r \cdot r = R$:  
        \[
        250v + 100(50,000 - v) = 6,500,000
        \]
\textbf{Step 3:} Simplify:  
        \[
        250v + 5,000,000 - 100v = 6,500,000
        \]
\textbf{Step 4:} Solve for $v$:  
        \[
        v = \frac{1,500,000}{150} = 10,000
        \]
\end{tcolorbox}

\begin{enumerate}[leftmargin=*]

\item[6.] \textbf{Derivation of Final Answer}: The final answer is then derived solely based on the logical reasoning applied. In this case: 

 \end{enumerate}
 
\begin{tcolorbox}[colback=white,colframe=customisedblue]
The number of VIP tickets sold is 10,000.
\end{tcolorbox}

\begin{figure}[t]
    \centering
    \includegraphics[width=\linewidth]{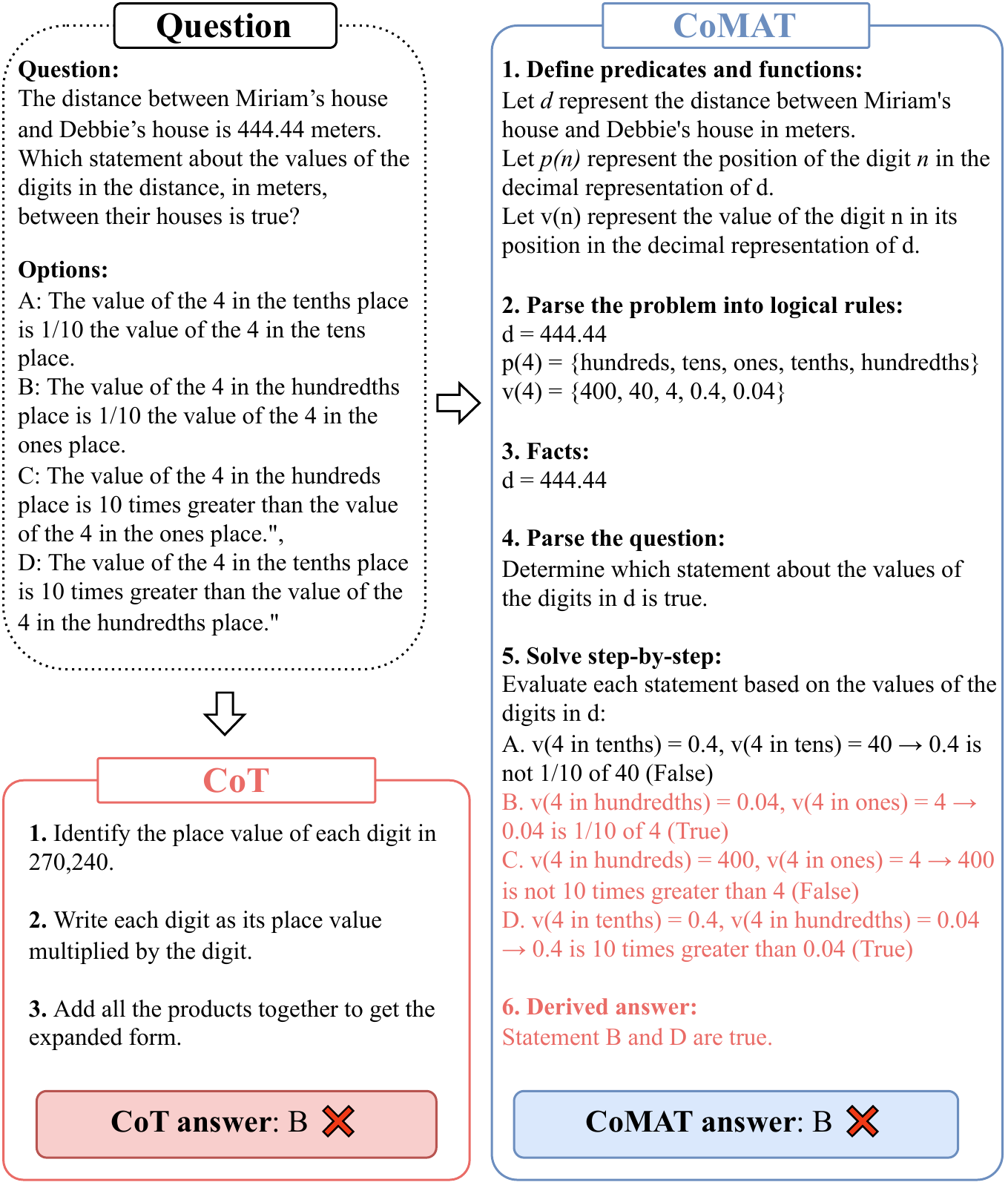}
    \caption{An example question from the MMLU Redux Elementary Mathematics dataset, comparing CoT and CoMAT. CoT follows a generic ``step-by-step'' approach without further guidance. In contrast, CoMAT enhances interpretability and verifiability by clearly pinpointing the error, which in this case arises from Step 5. %
    Traditional CoT, by comparison, lacks the ability to identify specific errors directly. %
    }
    \label{fig:error_comat}
\end{figure}

\begin{enumerate}[leftmargin=*]

\item[7.] (optional) \textbf{Answer Matching}: In multiple-choice QA tasks, CoMAT matches the final answer to the most similar one from the provided options, without considering the order or labelling of the options. 

\end{enumerate}
This step-by-step symbolic reasoning ensures that each operation is transparent and verifiable, reducing the potential for errors introduced in purely natural language reasoning. CoMAT's structured methodology facilitates easier error tracing, as shown in Figure~\ref{fig:error_comat}. In cases where the final answer is incorrect, each step can be individually examined to identify mistakes, an advantage over traditional CoT, where explanations may not align with the actual reasoning steps, making the reasoning process less precise or harder to verify~\cite{li2024towards}. The prompt for CoMAT, applying $S_{1} \dots S_{7}$, can be found in Appendix~\S\ref{sec:exactprompt}.

\section{Experimental Setup}
\paragraph{Datasets.}
We select a total of seven datasets, including five in English: AQUA~\cite{ling2017program}, MultiArith~\cite{roy2016solving} and GSM8K~\cite{cobbe2021training}, which were extracted from Math Word Problem (MWP) datasets, MMLU-Redux (\citealt{gema2024we}); focusing on various mathematics subjects, including abstract algebra, college-level mathematics, high school mathematics, and elementary mathematics,
Olympiad Bench (English; OlympBench-EN -- text-only; \citealt{he2024olympiadbench}), focused on text-based olympiad-mathematics problems, and another two Mandarin datasets: GaoKao (MCQ; \citealt{zhang2023evaluating}), Olympiad Bench (Chinese; OlympBench-CN -- text-only; \citealt{he2024olympiadbench}).
These datasets include multiple-choice questions (AQUA, MMLU-Redux, GaoKao MCQ) and string-valued answers (GSM8K, MultiArith, Olympiad Bench).
The inclusion of complex datasets like GaoKao and Olympiad Bench allows us to assess model performance on academically challenging, non-standard, questions, such as those in Olympiad-level exams and specialised academic tasks. %

\paragraph{Evaluation Metrics.} For multiple-choice datasets, we use exact match metrics, requiring the predicted answer to fully match the correct one, not just the option code (e.g., \textit{A}, \textit{B}). This ensures evaluation based on complete responses, addressing clarity concerns in datasets like MMLU-Redux. For string-valued answers, we also use an exact match in GSM8K. On Olympiad Bench, we use GPT-4o-mini as a benchmark to evaluate how well the model's answers align with the ground truth. Further details about GPT-4o-mini evaluation can be found in Appendix~\ref{sec: Implementation Details}.

\paragraph{Language Models.} 
We conduct our experiments using GPT-4o~\cite{achiam2023gpt}, Gemini-1.5-Pro~\cite{team2023gemini}, and Qwen2~\cite{yang2024qwen2}. This variety allows us to examine performance across different model architectures, sizes, and language proficiencies. For Faithful CoT, we used GPT-4, following the baselines from~\citet{lyu2023faithful}, as GPT-4o tends to produce more invalid outputs compared to GPT-4.

\paragraph{Baselines.} We compare our method against two baselines using the same decoding strategies: greedy decoding, where the most probable next token is selected, with a zero-shot setting. The baselines include: (1) Standard Prompting, using a similar prompt to that in Holistic Evaluation of Language Models~\cite{liang2022holistic}; (2) CoT prompting; %
(3) For common benchmarks, we additionally compare against \emph{Faithful CoT~\cite{lyu2023faithful}}, as it involves converting questions into symbolic representations and executing them in Python, making it a relevant baseline relative to existing Python-based methods~\cite{chen2022program, gao2023pal}. When results for specific tasks are unavailable, we reproduce the baseline results to ensure consistency and fairness in comparison. Verification solvers such as Datalog, Z3, and LEAN are excluded in $S_{5}$ due to their inability to directly solve problems, ensuring a focused and reliable comparison~\cite{olausson2023linc, gou2023critic}. The specific prompt for CoMAT can be found in Appendix~\S\ref{sec:exactprompt}.

\begin{table*}[t]
\centering
\resizebox{\textwidth}{!}{
\begin{tabular}{lccccccccc}
\toprule
\multirow{2}{*}{\textbf{Model}} & \multicolumn{3}{c}{\textbf{MWP}} & \multicolumn{3}{c}{\textbf{English}} & \multicolumn{3}{c}{\textbf{Chinese}} \\
\cmidrule(lr){2-4} \cmidrule(lr){5-7} \cmidrule(lr){8-10}
& \textbf{AQUA} & \textbf{GSM8K} & \textbf{MultiArith} & \textbf{MMLU-Redux} & \textbf{OlympBench} & \textbf{Average (EN+MWP)} & \textbf{GaoKao} & \textbf{OlympBench} & \textbf{Average (CN)} \\
\midrule
\multicolumn{10}{c}{\textit{Open Source Models}} \\
\midrule
Qwen2-7b & 34.25\% & 19.29\% & 47.78\% & 56.52\% & 5.19\% & 32.61\%\% & \textbf{48.98\%} & 8.09\% & 28.54\% \\
+ CoT & 33.07\% & \textbf{81.78\%} & \textbf{97.78\%} & 64.80\% & 3.86\% & 56.26\% & 40.82\% & 4.17\% & 22.50\% \\
+ CoMAT (Ours) & \textbf{42.13\%} & 79.80\% & 94.46\% & \textbf{79.80\%} & \textbf{20.92\%} & \textbf{57.42\%} & 44.90\% & \textbf{13.24\%} & \textbf{29.07\%} \\
\midrule
Qwen2-72b & 51.97\% & 37.43\% & 95.56\%& 66.53\% & 10.39\% & 52.38\% & 53.06\% & 11.27\% & 32.17\% \\
+ CoT & \textbf{79.13\%} & 82.76\% & 97.22\% & 79.17\% & 27.74\% & 73.20\% & 55.10\% & \textbf{19.66\%} & 37.38\% \\
+ CoMAT (Ours) & 72.44\% & \textbf{83.90\%} & \textbf{100.00\%} & \textbf{81.72\%} & \textbf{32.17\%} & \textbf{74.05\%} & \textbf{59.18\%} & 18.87\% & \textbf{39.03\%} \\
\midrule
\multicolumn{10}{c}{\textit{Closed Source Models}} \\
\midrule
Gemini & 48.03\% & 45.71\% & 97.22\% & 68.00\% & 5.50\% & 52.89\% & 43.00\% & 14.95\% & 28.98\% \\
+ CoT & 75.20\% & \textbf{90.51\%} & 98.33\% & 79.55\% & 21.28\% & 72.97\% & 65.31\% & 13.27\% & 39.29\% \\
+ CoMAT (Ours) & \textbf{78.74\%} & 90.43\% & \textbf{98.89\%} & \textbf{79.71\%} & \textbf{29.46\%} & \textbf{75.458\%} & \textbf{67.30\%} & \textbf{15.95\%} & \textbf{41.63\%} \\
\midrule
GPT-4o & 44.49\% & 56.72\% & \textbf{100.00\%} & 59.70\% & 9.94\% & 54.17\% & 36.73\% & 8.82\% & 22.78\% \\
+ CoT & \textbf{84.25\%} & \textbf{94.46\%} & \textbf{100.00\%} & 88.10\% & \textbf{41.84\%} & \textbf{81.73\%} & 63.27\% & 23.53\% & 43.40\% \\
+ CoMAT (Ours) & 83.46\% & 93.70\% & \textbf{100.00\%} & \textbf{88.30\%} & 40.42\% & 81.18\% & \textbf{71.43\%} & \textbf{26.47\%} & \textbf{48.95\%} \\
\midrule
\multicolumn{10}{c}{\textit{Additional Benchmarks}} \\
\midrule
Faithful CoT & 73.6\% & \textbf{95.0\%} & 99.2\% & 76.88\% & 0 & 0 & 0 & 0 & 0 \\
\bottomrule
\end{tabular}
}
\caption{We compare model performance across benchmarks, categorised by language (English and Chinese). 
}
\label{tab:results}
\end{table*}

\section{Results}

\begin{figure}[t]
    \centering
    \includegraphics[width=\linewidth]{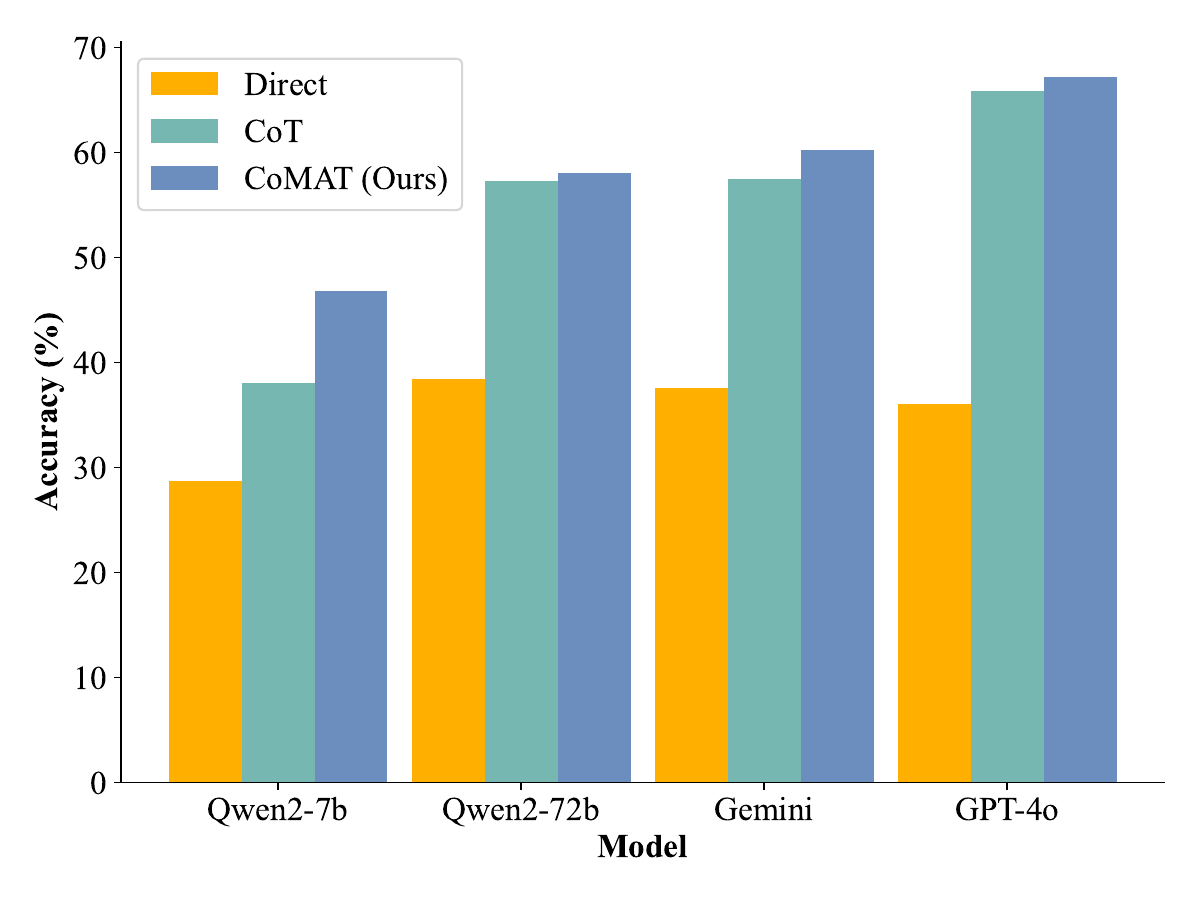}
    \caption{Average performance across all datasets for each model.}
    \label{fig:average_results_comparison}
\end{figure}

Table~\ref{tab:results} compares CoMAT, CoT~\cite{kojima2022large}, Faithful CoT~\cite{lyu2023faithful} and baseline models across various benchmarks. 
CoMAT outperforms traditional CoT and Faithful CoT in most datasets, particularly on tasks requiring advanced mathematical reasoning.

For open-source models, \textbf{Qwen2-7b} shows significant improvements with CoMAT. Under CoT, it often struggles to provide reasoning on the OlympiadBench dataset. With CoMAT, its performance increases on English OlympiadBench (5.19\% → 20.92\%), and Chinese OlympiadBench (8.09\% → 13.24\%).
\textbf{Qwen2-72b} also benefits from CoMAT, especially on OlympiadBench and GaoKao, with English OlympiadBench accuracy increasing (27.74\% → 32.17\%), 
highlighting CoMAT’s ability to improve reasoning in models that initially underperform.

Among closed-source models, \textbf{Gemini-1.5-Pro} shows notable gains with CoMAT, with an 8.18\% improvement on English OlympiadBench. Similarly, \textbf{GPT-4o} shows substantial gains on Mandarin datasets, with performance on GaoKao rising from 63.27\% (CoT) to 71.43\% (CoMAT). However, there are some cases where CoMAT does not outperform CoT. For instance, \textbf{Qwen2-72b} and \textbf{GPT-4o} show declines on AQUA (79.13\% → 72.44\% and 84.25\% → 83.46\%, respectively). This suggests that CoMAT may sometimes generate false conversions and outputs, as shown in Appendix~\ref{sec:Manual Annotation for CoMAT Verification}. These decreases suggest that on simpler tasks where models already perform well, the added complexity of symbolic reasoning may not yield significant benefits.

When comparing CoMAT with Faithful CoT on MWP datasets, Faithful CoT shows a minor gain on GSM8K (93.70\% → 95.0\%). However, CoMAT demonstrates a significant 9.86\% improvement on AQUA, highlighting its ability to outperform Faithful CoT without relying on external solvers. CoMAT also surpasses Faithful CoT on MMLU-Redux. Faithful CoT is excluded from GaoKao and OlympiadBench due to its inability to execute most questions using external solvers 
(Appendix~\ref{Faithful CoT vs CoMAT}). CoMAT effectively mitigates these limitations, proving its capability to handle complex tasks without reliance on external solvers.

We observe the most significant gains on challenging datasets like \textbf{GaoKao} and \textbf{OlympiadBench}, which require advanced reasoning. For example, CoMAT improves average performance on the English OlympiadBench (23.68\% → 30.74\%) and GaoKao (56.13\% → 60.70\%). On simpler datasets, gains are smaller but present, such as a 4.48\% increase on MMLU-Redux. Average results for each model are illustrated in Figure~\ref{fig:average_results_comparison}. 

\begin{table}[t]
\centering
\resizebox{\columnwidth}{!}{%
\begin{tabular}{lrrrrrr}
\toprule
\textbf{Subjects} & \textbf{L1} & \textbf{L2} & \textbf{L3} & \textbf{L4} & \textbf{L5} & \textbf{Total} \\
\midrule
Precalculus            & 33.30\% & 46.20\% & 13.30\% & 7.70\%  & 8.30\%  & 19.60\% \\
Inter. Algebra   & 42.90\% & 41.70\% & 73.70\% & 30.40\% & 16.70\% & 36.10\% \\
Algebra                & 88.20\% & 90.50\% & 65.40\% & 76.70\% & 40.00\% & 69.40\% \\
Number Theory          & 100.00\% & 60.00\% & 68.80\% & 68.40\% & 41.70\% & 64.50\% \\
Prealgebra             & 85.70\% & 68.40\% & 82.40\% & 55.00\% & 31.60\% & 61.00\% \\
Geometry               & 100.00\% & 75.00\% & 62.50\% & 40.00\% & 7.69\%  & 43.90\% \\
Counting \& Prob. & 50.00\% & 42.90\% & 50.00\% & 46.20\% & 33.30\% & 42.10\% \\
\bottomrule
\end{tabular}
}
\caption{Evaluation of Qwen2-7B on MATH500, decomposed into various subject domains and difficulty levels (Inter. Algebra is Intermedia Algebra; Prob. is Probability).}
\label{tab:math500-results}
\end{table}

We further analyse Qwen2-7B on the MATH500 benchmark~\cite{hendrycks2021measuring}, decomposing performance across domains and difficulty levels to understand further about the strengths and weaknesses of CoMAT towards the strengths and limitations of CoMAT in mathematical reasoning. As shown in Table~\ref{tab:math500-results}, the model performs strongly at levels 1--3, but its accuracy declines at levels 4 and 5 as complexity increases. The most pronounced weaknesses appear in precalculus, intermediate algebra, and counting \& probability at higher levels. However, it is relatively strong in pre-algebra, algebra, and number theory. These findings, evaluated using Qwen2-7B, highlight that CoMAT can yield competitive results even with smaller-scale models.

Overall, integrating symbolic reasoning into the Chain-of-Thought process significantly enhances language models' ability to tackle complex mathematical reasoning. This integration proves particularly valuable as complex reasoning typically benefits more from question decomposition through symbolic conversion, enabling the model to reduce ambiguity, rather than relying solely on natural language processing. While standard evaluation with CoMAT incurs a $1.5\times$ computational overhead, leveraging additional test-time computation 
~\cite{snell2024scaling}, using vLLM~\cite{kwon2023efficient} achieves comparable optimisation with only minor additional cost, as detailed in Appendix~\ref{sec:computational-cost}. The improvements are particularly notable on advanced tasks like \textbf{GaoKao} and \textbf{OlympiadBench}, showcasing CoMAT's strength in handling intricate reasoning through structured symbolic representations. While CoMAT performs strongly across various benchmarks, its true strength lies in solving unfamiliar and complex datasets.

\subsection{Step Contribution Analysis}
We conducted an ablation study to assess the impact of each step in the CoMAT prompt on model performance. We tested 16 variations by removing Steps 1, 2, 3, or 4 individually, as well as combinations (\ie removing steps 1 and 2).

\begin{figure}[t]
    \centering
    \includegraphics[width=\linewidth]{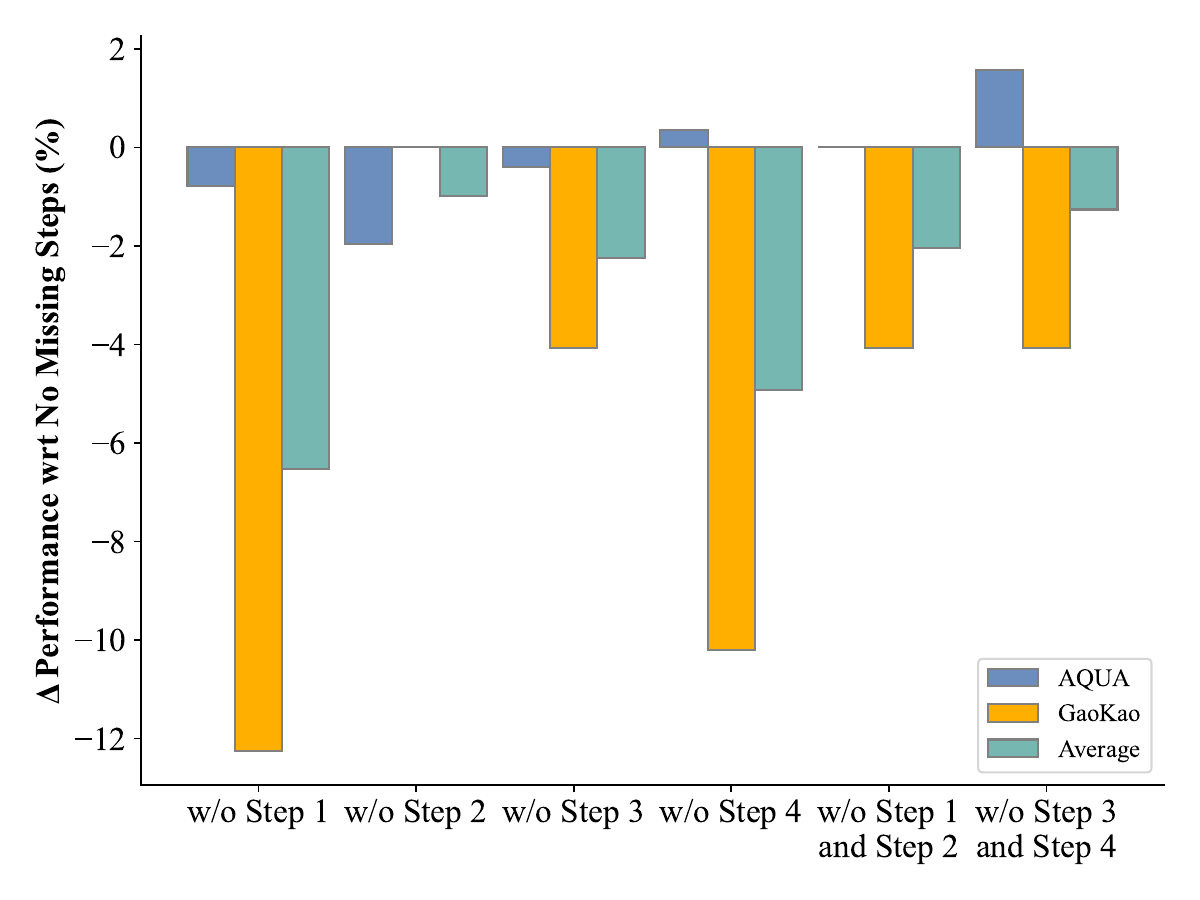}
    \caption{Performance change $(\Delta)$ for each configuration with missing steps.     Detailed results for all complete variants are provided in Appendix \ref{Detailed Results for Missing Steps}.}
    \label{fig:missing_steps_performances}
\end{figure}

Figure~\ref{fig:missing_steps_performances} shows the change in accuracy compared to the full CoMAT prompt. We found that each individual step plays a crucial role in maintaining the accuracy of CoMAT. Removing Step 1 leads to the most significant drop in accuracy (6.91\%). This suggests that Step 1 is fundamental to CoMAT, potentially serving as the foundation upon which other steps build; its absence may disrupt critical initial processes or data preparation necessary for accurate performance. Omitting Step 2 or Step 3 results in smaller declines (1.38\% and 2.63\%, respectively), indicating these steps are important but less critical than Step 1. Interestingly, removing both Steps 1 and 2 results in a smaller performance drop (2.43\%) than removing Step 1 alone, suggesting overlapping functionalities where the model compensates using remaining steps. Overall, the full CoMAT prompt achieves the highest average accuracy of 77.45\%, reaffirming the importance of retaining all steps for optimal performance.

\begin{figure}[t]
    \centering
    \begin{subfigure}[b]{0.48\linewidth}
        \includegraphics[width=\linewidth]{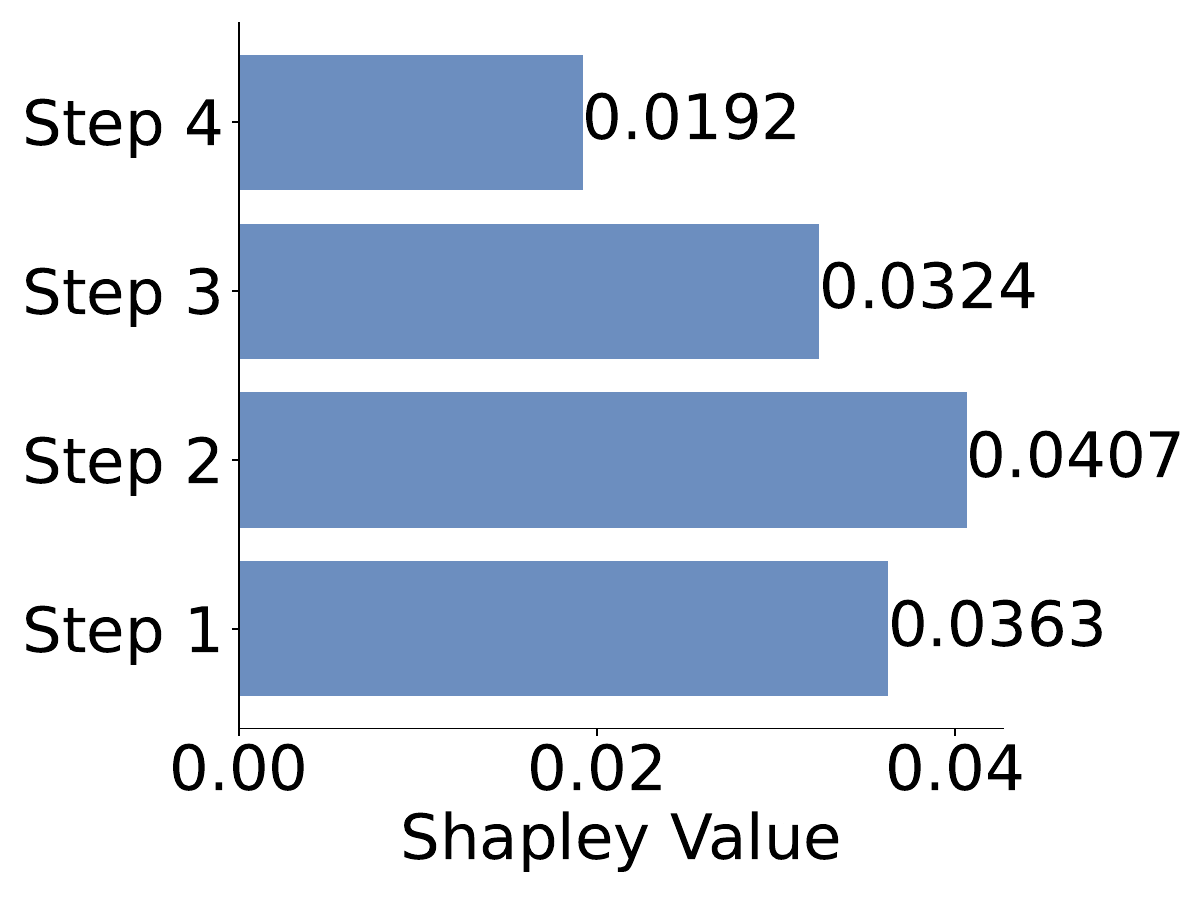}
        \caption{Shapley values for each step.} \label{fig:shapley_values_bar_plot}
    \end{subfigure}
    \hfill
    \begin{subfigure}[b]{0.48\linewidth}
        \includegraphics[width=\linewidth]{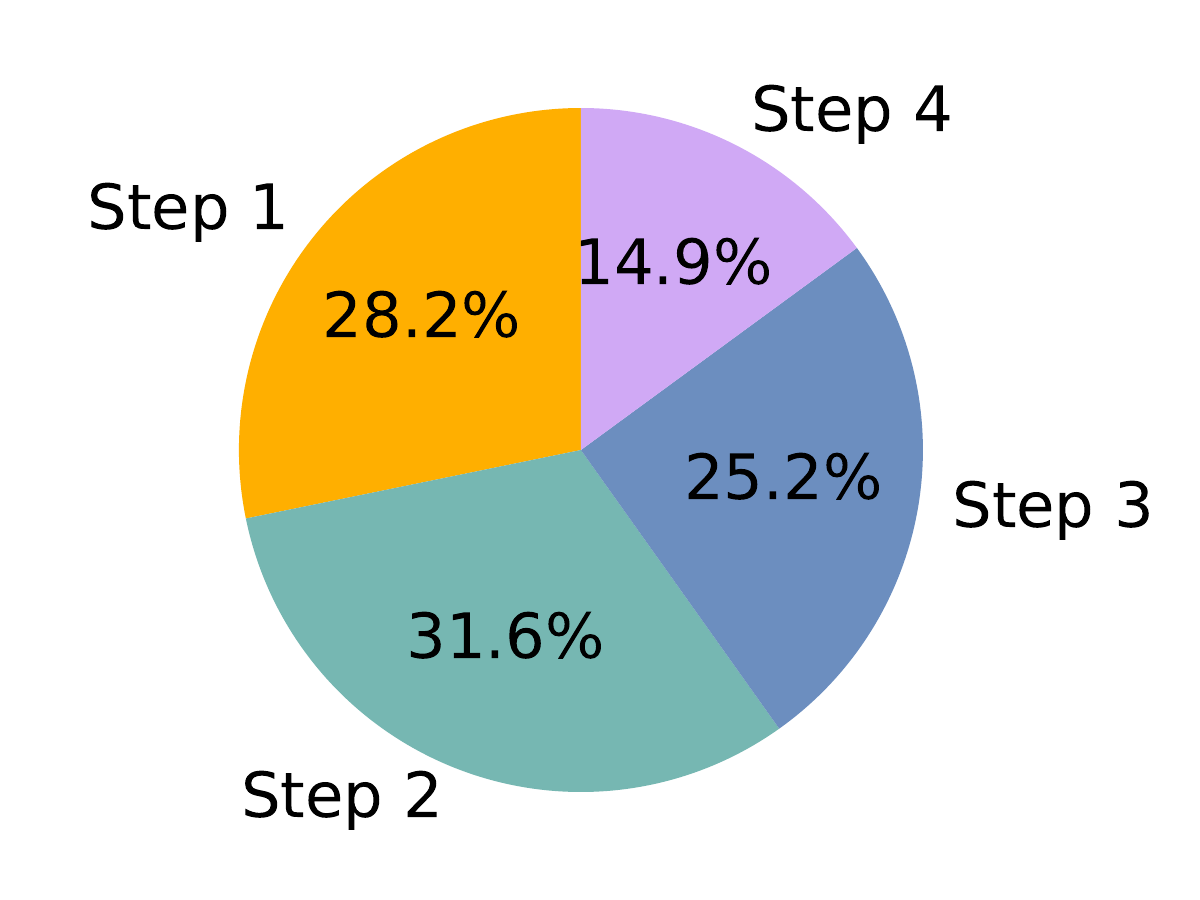}
        \caption{Percentage impact of steps.} \label{fig:shapley_values_pie_chart}
    \end{subfigure}
    \caption{(a) and (b) represent the Shapley values in bar and pie chart form, respectively.}
    \label{fig:shapley_values_comparison}
\end{figure}

We further quantified each step's contribution using Shapley-value \cite{shapley1953value} analysis (Figure~\ref{fig:shapley_values_comparison}). The analysis shows that all steps positively impact performance, with Steps 1 and 2 having the greatest influence, aligning with the expectation that initial steps lay the foundation for subsequent reasoning. For a detailed breakdown of the Shapley analysis calculations, refer to Appendix~\ref{sec:SHAP Analysis Experimental Detail}.

These results highlight that every step in CoMAT’s pipeline is essential, not only for achieving high accuracy but also for enhancing interpretability and verifiability. Each step builds upon the previous ones, ensuring that the reasoning process remains transparent and coherent. This reinforces CoMAT’s strength in balancing empirical performance with faithful, interpretable reasoning, making it a robust and reliable framework across diverse mathematical reasoning tasks.

\subsection{Multilingual Analysis}
To assess the impact of low-resource languages on CoMAT’s performance, we evaluate the models on the MGSM dataset~\cite{shi2022language}, focusing on Swahili (SW), Bengali (BN), Thai (TH), and Telugu (TE). While CoMAT has shown strong performance in high-resource languages like English and Mandarin, our analysis reveals mixed results in low-resource language settings relative to CoT. 

As shown in Table~\ref{tab:mgsm_results},
Gemini-1.5-Pro demonstrated notable improvements in low-resource contexts, with an average performance increase of 5.60\% after applying CoMAT. Significant gains were observed in Bengali (BN) and Swahili (SW), where accuracy increased to 82.40\% and 83.20\%, respectively. These improvements indicate that CoMAT can enhance reasoning in lower-resource settings when applied to certain models.

\begin{table}[t]
\centering
\scriptsize
\begin{tabular}{lccccc}
\toprule
\textbf{Model} & \textbf{BN} & \textbf{SW} & \textbf{TE} & \textbf{TH} & \textbf{Average} \\
\midrule
\multicolumn{6}{c}{\textit{GPT-4o Models}} \\
\midrule
non-CoT      & 52.80 & 52.40 & 51.60 & 58.00 & 53.70\%\\
+CoT          & 92.00 & 88.80 & 86.00 & 91.20 & \textbf{89.50\%}\\
+CoMAT        & 89.60 & 88.00 & 84.80 & 90.80 & 88.30\%\\
\midrule
\multicolumn{6}{c}{\textit{Gemini Models}} \\
\midrule
non-CoT      & 40.00 & 46.00 & 41.60 & 42.00 & 42.40\% \\
+CoT          & 79.20 & 70.80 & 62.40 & 84.00 & 74.10\%\\
+CoMAT        & 82.40 & 83.20 & 74.00 & 79.20 & \textbf{79.70\%}\\
\bottomrule
\end{tabular}
\caption{Performance comparison of GPT-4o and Gemini models across low-resource MGSM benchmarks (BN, SW, TE, TH).}
\label{tab:mgsm_results}
\end{table}

However, the performance of the GPT-4o model slightly declines, from 89.50\% to 88.30\%. This suggests that while CoMAT remains effective overall, certain models might struggle to generalise as effectively in low-resource contexts. In languages like Thai (TH) and Telugu (TE), the model’s performance remained relatively stable but did not show the same level of improvement observed in higher-resource languages. These indicate that while CoMAT shows promise for improving reasoning in low-resource languages, particularly with models like Gemini-1.5-Pro, further optimization may be required to optimise its performance across models and language contexts with limited training data.

\subsection{Answer Order Swapping}

The robustness of the CoMAT methodology lies in its reliance on symbolic representations, which create a structured and uniform framework for problem-solving, reducing ambiguity and variability. This makes the model less sensitive to changes in dataset order, linguistic nuances, or variations in answer choices. Building on the findings of \citet{gupta2024changing}, which demonstrated that altering answer choices in the MMLU dataset could affect model accuracy, we extended this investigation to both the MMLU-Redux and AQUA datasets using GPT-4o as our baseline. We have also introduced an additional challenge by including a random answer option, as detailed in Appendix~\ref{sec:option-swapping}.\\

Figure~\ref{fig:aqua_mmlu_comparison}
illustrates the effect of this option swapping. On the AQUA dataset, both CoT and CoMAT models experienced accuracy drops, highlighting their sensitivity to answer structure changes. However, CoMAT was more resilient, with only a 0.17\% decrease and a low standard deviation (s.d.) of 0.91\%. In contrast, GPT-4o with CoT saw a significant drop of 14.96\% and a higher s.d. of 3.50\%, indicating greater inconsistency when options were shuffled. This highlights CoMAT’s stability and robustness under altered conditions.

On the MMLU-Redux dataset, GPT-4o baseline showed a slight improvement, with a s.d. of 1.34\%. CoT and CoMAT both had minor accuracy decreases, with CoMAT achieving 85.05\% (s.d. of 1.65\%), slightly outperforming CoT at 84.43\% (1.87\% s.d.). Detailed results can be referred to in Appendix~\ref{Detailed Results for Missing Steps}. These results demonstrate CoMAT’s consistency when confronted with the complexities of option swapping. Other methods exhibit greater variability and performance sensitivity.

\begin{figure}
    \centering
    \includegraphics[width=\linewidth]{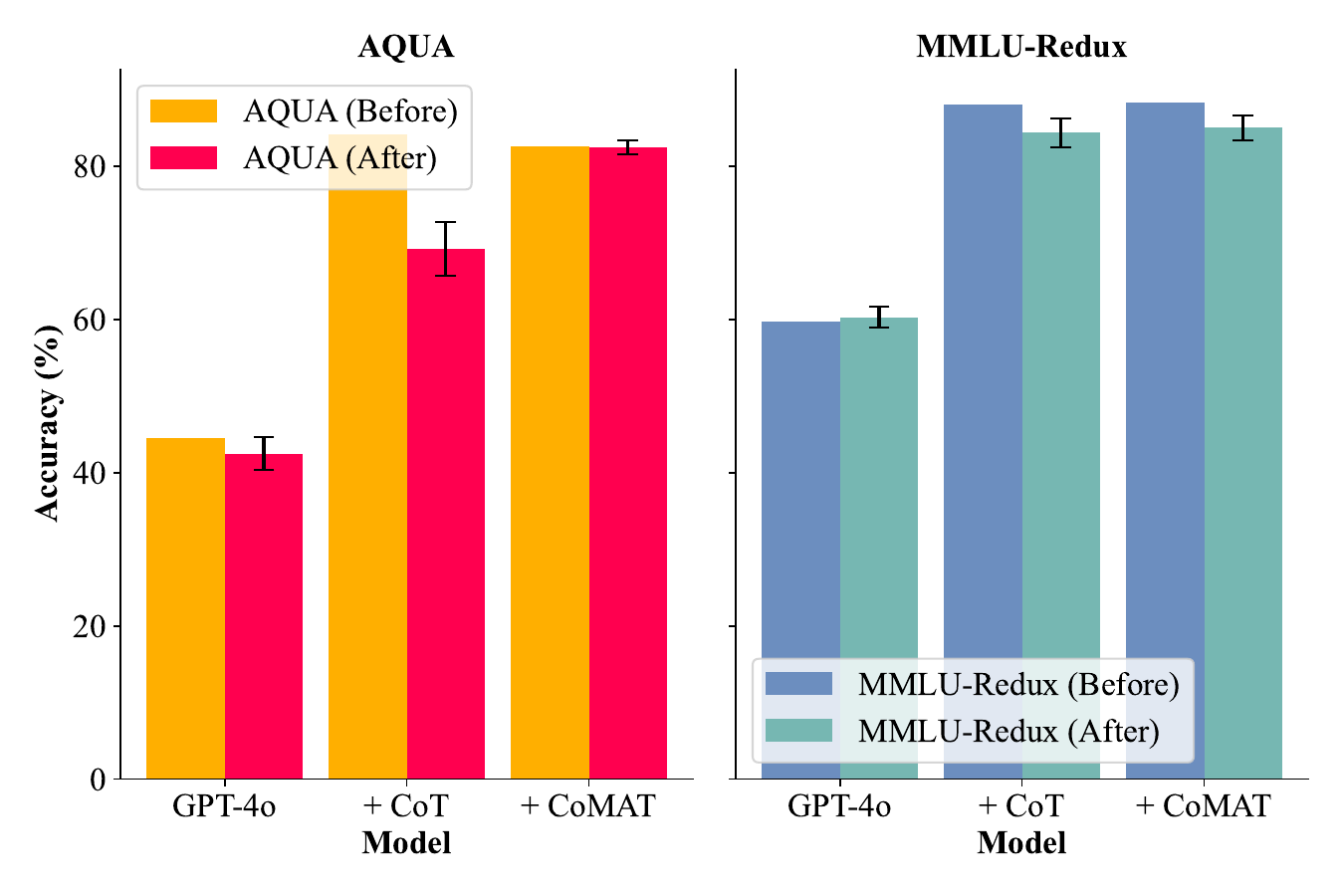}
    \caption{Average accuracy for AQUA and MMLU before and after option changing.} \label{fig:aqua_mmlu_comparison}
\end{figure}

\section{Related Work}

\textbf{Logical Reasoning.} Logical reasoning tasks require models to handle complex logical structures~\cite{cummins1991conditional}. Traditional methods include rule-based~\cite{robinson1965machine} and neural methods~\cite{amayuelas2022neural, gerasimova2023comparative} for interpreting symbolic representations. Recent methods, such as Logic-LM~\cite{pan2023logic}, SAT-LM~\cite{ye2024satlm}, and LeanReasoner~\cite{jiang2024leanreasoner}, use LLMs to convert natural language into symbolic syntax, processed by external tools. These frameworks enhance performance through self-consistency and non-linear reasoning~\cite{wang2023plan, zhang2022automatic} but assume LLMs are less reliable than rule-based reasoners for parsing symbolic expressions.

\textbf{Symbolic Chain-of-Thought Prompting.} Symbolic CoT prompting~\cite{lyu2023faithful} integrates natural language (NL) and symbolic language (SL) in reasoning, with NL breaking queries into subproblems and SL programs (e.g., Python) solving them. 
Recent methods~\cite{ li2023symbolic} such as Faithful Logical Reasoning via Symbolic CoT reduce SL reliance by leveraging LLMs in symbolic reasoning~\cite{xu2024faithful}. However, these methods primarily address domains such as logical reasoning rather than mathematical reasoning. Our work, concurrent with~\cite{arakelyan2024flare}, instead focuses on mathematical reasoning.

\noindent \textbf{Mathematical Reasoning.} Mathematical reasoning with LLMs has been explored widely~\cite{lewkowycz2022solving, luo2024improve, ahn2024large, imani2023mathprompter, chen2024masked, meadows2022survey, mirzadeh2024gsm}, with CoT methods yielding significant performance gains~\cite{jiang2024llms, chu2023survey, ranaldi2024aligning, leang2025picsar}. Deep problem understanding~\cite{zhong2024achieving}, structured formats~\cite{tam2024let}, and building supervision models for reasoning~\cite{lightman2023let, jiang2024rationalystpretrainingprocesssupervisionimproving} also enhance accuracy. Other studies focus on premise selection and symbolic frameworks for systematic evaluation~\cite{meadows2023symbolic, ferreira2020premise}. Recently, CoMAT has also been applied to improve the autoformalisation of LLMs in the LEAN theorem prover~\cite{leang2025theorem}.

\section{Conclusion}

We propose CoMAT, a simple yet effective framework that decomposes complex mathematical reasoning into two stages:  Symbolic Conversion and Reasoning Execution. CoMAT operates entirely within LLMs, eliminating reliance on external solvers and ensuring a transparent and accurate reasoning process. By avoiding external solvers, CoMAT mitigates issues related to code generation failures, providing a more robust solution for a broad range of mathematical tasks. Our analysis highlights four key steps in the CoMAT process and demonstrates its effectiveness across various datasets with different levels of complexity and linguistic diversity, including English, Mandarin, and low-resource languages. CoMAT consistently outperforms traditional CoT on the majority of selected English datasets, as well as on the average across selected Mandarin datasets. It also enhances consistency when answer options are shuffled, demonstrating both robustness and reliability. Despite its simplicity, CoMAT offers a scalable and effective solution for complex mathematical reasoning, providing greater faithfulness and verifiability across a wide range of tasks.

\section*{Limitations}
While CoMAT demonstrates strong performance, there are several potential limitations. One key challenge lies in the symbolic conversion process, which involves four steps. It remains difficult to automatically check and correct errors within these steps, often requiring manual annotation. Addressing this limitation presents an important avenue for future research. Secondly, our evaluation was limited to the current set of symbolic steps, where some steps have been previously used in other reasoning frameworks. For instance, a step similar to our step 1 is used by~\newcite{lyu2023faithful} and \newcite{xu2024faithful}. Although our approach proved effective, further research is needed to assess CoMAT's performance using additional symbolic languages to ensure a more comprehensive evaluation. Secondly, while CoMAT enhances verifiability and faithfulness, it introduces a higher computational overhead compared to CoT due to the structured nature of its formalisations. This process involves generating additional symbolic representations, leading to a larger token count, which increases both computational costs and API usage. \emph{Thirdly, accurately quantifying conversions, symbolic reasoning, and faithfulness quality control within symbolic conversions remains challenging without resource-intensive manual annotation, constituting a significant open research direction}~\cite{yang2024formal}. In our work, we propose that our overall accuracy improvements serve as an indirect metric for conversion quality. Consequently, scaling CoMAT could require more substantial computational resources than traditional CoT methods. Lastly, CoMAT focuses on mathematical reasoning. Further research could explore extending symbolic reasoning to other domains, beyond mathematical reasoning, to evaluate its effectiveness in broader tasks.

\section*{Acknowledgements}

We are grateful for the feedback from Dongwei Jiang, Yftah Ziser, Emile Van Krieken, Yu Zhao,  Alessio Devoto, Rohit Saxena, Giwon Hong, Xiaotang Du, and the reviewers that made the paper better. We are grateful for UKRI's support through the provision of computational resources at the University of Birmingham (Baskerville), the University of Bristol (Isambard AI), and the University of Edinburgh (Edinburgh International Data Facility).

\bibliography{anthology,custom}

\clearpage

\appendix

\section{Faithful CoT vs CoMAT}
\label{Faithful CoT vs CoMAT}

\begin{figure}[H]
\begin{tcolorbox}[
    colback=gray!25!white,
    colframe=gray!75!black,
    colbacktitle=gray!90!white,
    title=Verification of solutions - CoT (Faithful-CoT),
    fonttitle=\bfseries,
    width=\textwidth,
    boxrule=0.5pt,
    arc=4pt,
    auto outer arc
]
\small
\textbf{Question:} 

Xenia and Sergey play the following game. Xenia thinks of a positive integer \(N\) not exceeding 5000. Then she fixes 20 distinct positive integers \(a_{1}, a_{2}, \ldots, a_{20}\) such that, for each \(k=1,2, \ldots, 20\), the numbers \(N\) and \(a_{k}\) are congruent modulo \(k\). By a move, Sergey tells Xenia a set \(S\) of positive integers not exceeding 20, and she tells him back the set \(\{a_k: k \in S\}\) without spelling out which number corresponds to which index. How many moves does Sergey need to determine for sure the number Xenia thought of?

\textbf{Answer:} 2

\vspace{0.5cm}

\textbf{Faithful-CoT:}

\begin{enumerate}
    \item What is the maximum number Xenia can think of? \\
    Xenia thinks of a positive integer \(N\) not exceeding 5000.
\begin{itemize}
    \item max\_number = 5000
\end{itemize}

\begin{itemize}
    \item How many distinct positive integers does Xenia fix? \\
    She fixes 20 distinct positive integers \(a_{1}, a_{2}, \ldots, a_{20}\).
\end{itemize}
\begin{itemize}
 \item distinct\_integers = 20
\end{itemize}

\begin{itemize} 
    \item What is the condition for the numbers \(N\) and \(a_k\)? \\
    The numbers \(N\) and \(a_k\) are congruent modulo \(k\) for each \(k = 1, 2, \ldots, 20\).
\end{itemize}
\begin{itemize} 
 \item Condition: N \% k == a\_k \% k
\end{itemize}
    
    \item What is a move in the game? \\
    \begin{itemize}
    \item  Sergey tells Xenia a set \(S\) of positive integers not exceeding 20, and she tells him back the set \(\{a_k: k \in S\}\) without spelling out which number corresponds to which index.\\ Move:Sergey asks set S; Xenia gives ${a_k: k \in S}$
        \end{itemize}
    
    \item How many moves does Sergey need to determine for sure the number Xenia thought of? \\
    \begin{itemize}
    \item This requires finding the least common multiple of the numbers from 1 to 20 and figuring out how many times this number fits into the maximum number Xenia can think of. Sergey can then determine the number Xenia thought of by asking for the set \(\{a_k: k \in S\}\) for different sets \(S\) until he has enough information to determine \(N\).
    \end {itemize}
    \begin{lstlisting}[language=Python, basicstyle=\color{red}]
lcm_1_to_20 = np.lcm.reduce(range(1, 21))
possible_numbers = max_number // lcm_1_to_20
moves_needed = math.ceil(math.log2(possible_numbers))
    \end{lstlisting}
    \item Final Answer: How many moves does Sergey need to determine for sure the number Xenia thought of? \\
\textcolor{red}{\textbf{answer = moves\_needed}}
    
\end{enumerate}
\end{tcolorbox}
\end{figure}

\begin{figure*}[ht]
\begin{tcolorbox}[
    colback=gray!25!white,
    colframe=gray!75!black,
    colbacktitle=gray!90!white,
    title=Verification of solutions - CoT (CoMAT),
    fonttitle=\bfseries,
    width=\textwidth,
    boxrule=0.5pt,
    arc=4pt,
    auto outer arc
]
\small

\textbf{Question:} 

Xenia and Sergey play the following game. Xenia thinks of a positive integer \(N\) not exceeding 5000. Then she fixes 20 distinct positive integers \(a_{1}, a_{2}, \ldots, a_{20}\) such that, for each \(k=1,2, \ldots, 20\), the numbers \(N\) and \(a_{k}\) are congruent modulo \(k\). By a move, Sergey tells Xenia a set \(S\) of positive integers not exceeding 20, and she tells him back the set \(\{a_k: k \in S\}\) without spelling out which number corresponds to which index. How many moves does Sergey need to determine for sure the number Xenia thought of?

\textbf{Answer:} 2

\vspace{0.5cm}

\textbf{CoMAT:}

\begin{enumerate}
    \item Define predicates, functions, and variables: 
    \begin{itemize}
        \item Let \(N\) be the positive integer Xenia thinks of, where \(1 \leq N \leq 5000\).
        \item Let \(a_1, a_2, \ldots, a_{20}\) be 20 distinct positive integers such that \(N \equiv a_k \pmod{k}\) for each \(k = 1, 2, \ldots, 20\).
        \item Let \(S\) be a set of positive integers not exceeding 20 that Sergey chooses.
        \item Let \(A(S)\) be the set of integers Xenia returns, which is \(\{a_k: k \in S\}\).
    \end{itemize}
    
    \item Parse the problem into logical rules: 
    \begin{itemize}
        \item For each \(k\), \(N \equiv a_k \pmod{k}\).
        \item Sergey can choose a set \(S \subseteq \{1, 2, \ldots, 20\}\).
        \item Xenia returns the set \(\{a_k: k \in S\}\) without indicating which \(a_k\) corresponds to which \(k\).
    \end{itemize}
    
    \item Write all the facts explicitly mentioned:
    \begin{itemize}
        \item \(N\) is a positive integer such that \(1 \leq N \leq 5000\).
        \item \(a_1, a_2, \ldots, a_{20}\) are distinct and satisfy \(N \equiv a_k \pmod{k}\).
    \end{itemize}
    
    \item Parse the question into symbolic form:
    \begin{itemize}
        \item Determine the minimum number of moves Sergey needs to make to uniquely identify \(N\).
    \end{itemize}
    
    \item Solve step by step:
    \begin{itemize}
        \item The key to solving this problem is to understand that Sergey needs to gather enough information to determine \(N\) uniquely.
        \item Since \(N \equiv a_k \pmod{k}\) for each \(k\), the set of congruences forms a system of simultaneous congruences.
        \item By the Chinese Remainder Theorem (CRT), if the moduli are pairwise coprime, the system of congruences has a unique solution modulo the product of the moduli.
        \item The numbers 1 through 20 are not all pairwise coprime, but we can consider subsets of them that are.
        \item Sergey needs to choose sets \(S\) such that the product of the moduli in each set covers all possible residues for \(N\) modulo 1 through 20.
    \end{itemize}
    
    \item Derived answer:
    \begin{itemize}
        \item Sergey needs to make 2 moves to determine \(N\) for sure. In the first move, he can choose a set that gives him enough information to narrow down the possibilities significantly, and in the second move, he can pinpoint the exact \(N\).
    \end{itemize}
\end{enumerate}

\textbf{Final Answer:} 2

\end{tcolorbox}
\end{figure*}
\clearpage

\section{Implementation Details}
\label{sec: Implementation Details}
In all our experiments, we used OpenAI GPT-4o models (gpt-4o-2024-08-06 and gemini-1.5-pro-001) as well as Qwen2-7B and Qwen2-72B. For API details regarding GPT-4o, please refer to \url{https://platform.openai.com/docs/models/gpt-4o}, and gemini can be reffered at \url{https://ai.google.dev/gemini-api/docs/models/gemini}. 
For the open-source models, Qwen2-7B and Qwen2-72B (\url{https://huggingface.co/docs/transformers/en/model_doc/qwen2}) serve as our primary inference models, and we conducted all experiments using two 80-GB A100 GPUs per dataset for inference.
The following hyperparameters were consistently applied across all experiments:

\begin{itemize}[leftmargin=*]
    \item \textbf{Temperature:} 0.0 (for greedy decoding) 
    \item \textbf{Max tokens:} 3500 
    \item \textbf{Experiment Setting:} All experiments are conducted through a 1-shot setting, with the same example across all evaluations.
\end{itemize}

\subsection{Evaluation Metrics}
Majority of our datasets use exact match as our evaluation metrics. For the OlympiadBench datasets, where exact match is not a suitable evaluation metric, we used the GPT-4o-mini model as a benchmark to assess how closely the answers of the model aligned with the ground truth. To prevent the model from generating reasoning that could influence its decision, we input only the final three sentences, which typically contain both the answer and the correct solution. The task was to determine whether these two elements matched, without providing any intermediate reasoning. We would like to calrify that our approach differs from the traditional \textit{LLM-as-a-judge} evaluation, as we restrict the context to the final three sentences and assess only whether the predicted answer aligns with the ground truth, without evaluating the intermediate reasoning. The prompt used for this evaluation is presented below:

\begin{figure}[ht]
\begin{tcolorbox}[
    colback=gray!25!white,
    colframe=gray!75!black,
    colbacktitle=gray!90!white,
    title=GPT-4o Evaluation Prompt,
    fonttitle=\bfseries,
    width=\linewidth,
    boxrule=0.5pt,
    arc=4pt,
    auto outer arc,
]
\small
\textbf{System Message:} \\
You are a decider that decides whether the answer is the same as the correct answer. If the output doesn't align with the correct answer, respond with '0', whereas if it's correct, then respond with '1'. \textbf{DO NOT PROVIDE YOUR OWN ANSWER OR REASONING, JUST SELECT `0' OR `1'.} \\

\textbf{User Message:} \\
GPT-4o Result: \{gpt\_result\} \\
Correct Answer: \{correct\_answer\}. \\
Answer with 0 (Wrong) or 1 (Correct).
\end{tcolorbox}
\end{figure}

\section{Extended Results and Analysis}
In this section, we will present more results and analysis that did not fit into the main text:
\subsection{Detailed Results for Missing Steps}
\label{Detailed Results for Missing Steps}

We examine the sensitivity of various prompts in CoMAT by experimenting with 16 different variants. In each variant, we omit individual steps or combinations of steps from the CoMAT process to assess their impact on performance.

\begin{table}[t]
    \centering
    \resizebox{\linewidth}{!}{%
    \begin{tabular}{lccc}
        \hline
        \textbf{Missing Steps} & \textbf{AQUA (\%)} & \textbf{GaoKao (\%)} & \textbf{Average (\%)} \\
        \hline
        Missing Step 1 & 81.89 & 59.18 & 70.54 \\
        Missing Step 2 & 80.71 & 71.43 & 76.07 \\
        Missing Step 3 & 82.28 & 67.35 & 74.82 \\
        Missing Step 4 & 83.04 & 61.22 & 72.13 \\
        Missing Step 1, 2 & 82.68 & 67.35 & 75.02 \\
        Missing Step 3, 4 & 84.25 & 67.35 & 75.80 \\
        Missing Step 1, 3 & 83.07 & 69.39 & 76.23 \\
        Missing Step 2, 4 & 84.25 & 69.39 & 76.82 \\
        Missing Step 1, 4 & 82.28 & 65.31 & 73.80 \\
        Missing Step 2, 3 & 83.86 & 55.10 & 69.48 \\
        Missing Step 1, 2, 3 & 81.89 & 55.10 & 68.50 \\
        Missing Step 1, 3, 4 & 83.07 & 71.43 & 77.25 \\
        Missing Step 1, 2, 4 & 81.10 & 73.47 & 77.29 \\
        Missing Step 2, 3, 4 & 81.50 & 61.22 & 71.36 \\
        Missing Steps(CoT) & 84.25 & 63.27 & 73.76 \\
        No Missing Steps & 83.46 & 71.43 & \textbf{77.45} \\
        \hline
    \end{tabular}
    }
    \caption{Ablation study results that accompany Figure~\ref{fig:missing_steps_performances}. Performance comparison of different missing step configurations on AQUA and GaoKao datasets.}
    \label{tab:missing_steps_performance}
\end{table}

As shown in Table~\ref{tab:missing_steps_performance}, while AQUA and GaoKao occasionally perform better when certain steps are omitted, the overall average performance consistently decreases compared to the original CoMAT prompt. This indicates that CoMAT performs most effectively when all steps are included. The results highlight the importance of each step in contributing to the overall performance, demonstrating that CoMAT's structure is essential for achieving optimal results. 

\subsection{Option Swapping}
\label{sec:option-swapping}

\begin{figure*}[t]
    \centering
    \includegraphics[width=0.7\linewidth]{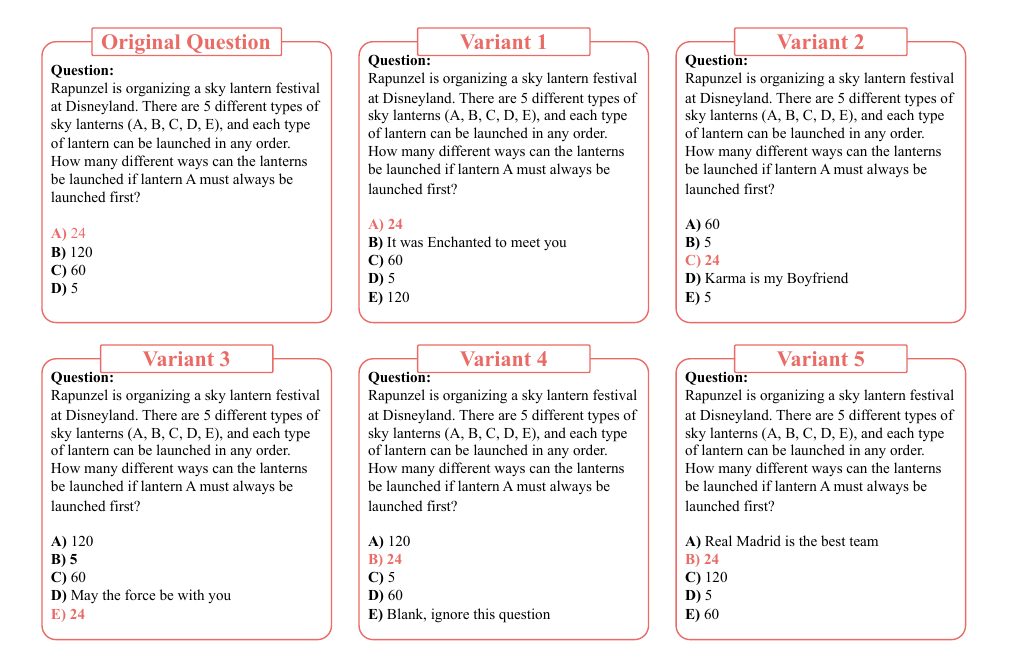}
    \caption{Multiple Variants of Option Swapping. This figure illustrates the variants after swapping all options and introducing five additional variants for comparison.}
    \label{fig:figures/shuffle_CoMAT}
\end{figure*}

This section supports the ablation studies presented in Figure~\ref{fig:aqua_mmlu_comparison}. We experimented by shuffling the options and adding an additional random option, creating 5 different variants for the choice sets. These variants are illustrated in Figure~\ref{fig:figures/shuffle_CoMAT}.

For each dataset, we conducted three evaluations per variant and calculated the average to ensure consistency. The results reveal a notable difference between CoT and CoMAT in terms of performance stability. CoMAT exhibits more consistent results following the option modifications, as evidenced by the standard deviation. For example, in the AQUA dataset, CoMAT demonstrated a low standard deviation of 0.91\%, whereas CoT had a significantly higher standard deviation of 3.50\%, underscoring the variability in CoT’s performance. These findings further support the consistency of CoMAT’s performance, as previously highlighted in Figure~\ref{fig:aqua_mmlu_comparison}.
Although both models showed a decline in performance after the option changes, CoMAT’s stability is worth noting. The drop in performance across models suggests that further investigation into the impact of option variation may be valuable.

\begin{figure*}[ht]
    \centering
    \begin{subfigure}[b]{\textwidth}
        \centering
        \includegraphics[width=0.8\linewidth]{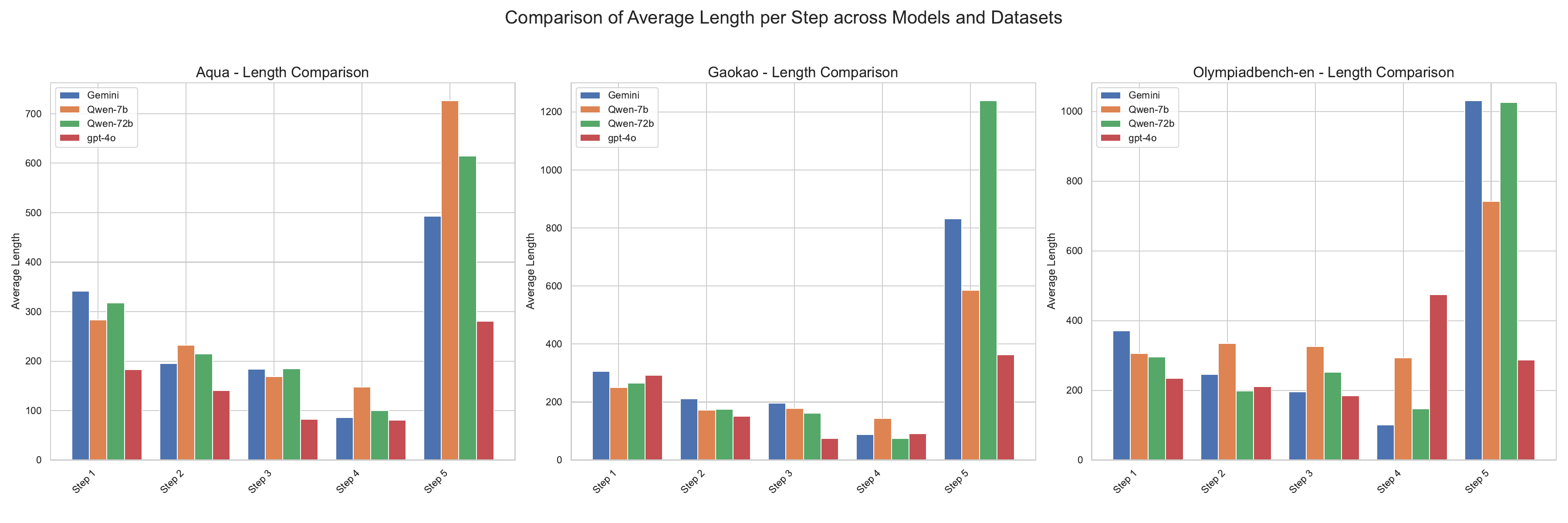}
        \caption{Word length comparison.}
        \label{fig:length_word_overview}
    \end{subfigure}
    
    \vspace{1em} %
    
    \begin{subfigure}[b]{\textwidth}
        \centering
        \includegraphics[width=0.8\linewidth]{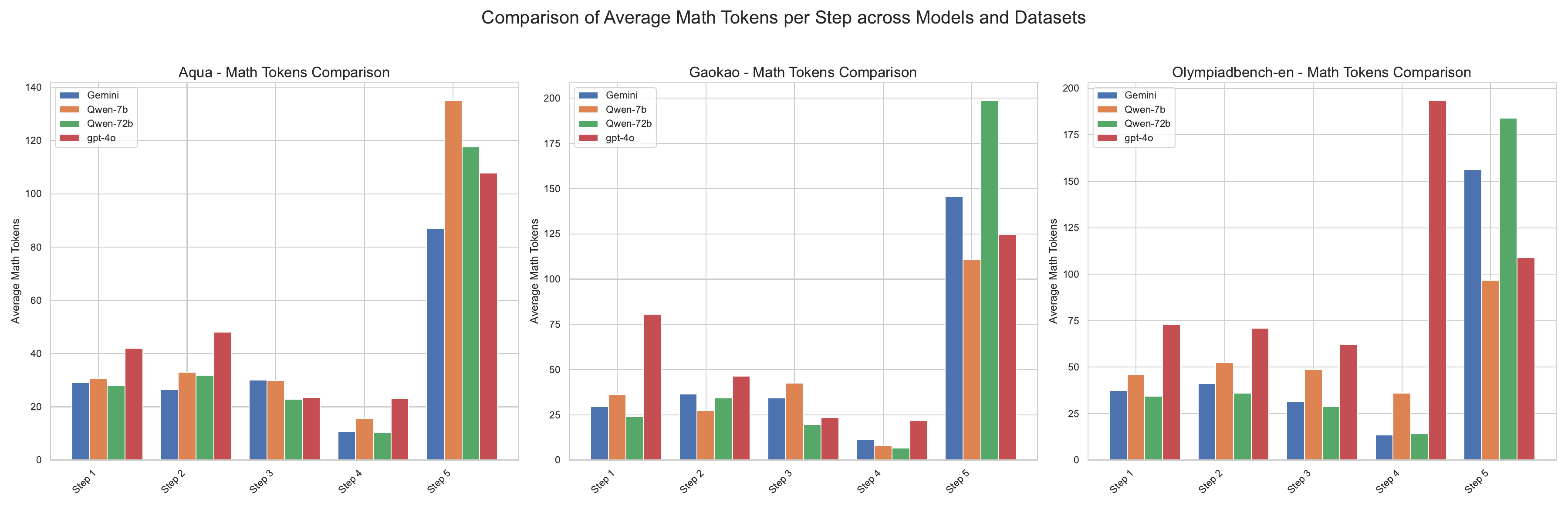}
        \caption{Math tokens count comparison.}
        \label{fig:math_word_overview}
    \end{subfigure}

    \caption{Comparison of average length and math tokens count.}
    \label{fig:comparison_length_math_tokens}
\end{figure*}

\begin{table}[htp]
    \centering
    \resizebox{\linewidth}{!}{%
    \renewcommand{\arraystretch}{1.2} %
    \begin{tabular}{lcccc}
        \hline
        \textbf{Swapping} & \textbf{CoMAT (\%)} & \textbf{CoT (\%)} & \textbf{non-CoT (\%)} \\
        \hline
        \multicolumn{4}{c}{\textbf{AQUA}} \\
        \hline
        Option Changing 1 & 83.07 & 73.23 & 44.49\\
        Option Changing 2 & 83.07 & 66.54 & 42.91\\
        Option Changing 3 & 81.49 & 68.11 & 40.16\\
        \hline
        \textbf{Average} & \textbf{82.54} & \textbf{69.29} & \textbf{42.52} \\
        \hline
        \multicolumn{4}{c}{\textbf{MMLU-Redux}} \\
        \hline
        Option Changing 1 & 86.94 & 86.57 & 58.79\\
        Option Changing 2 & 84.36 & 83.61 & 61.14\\
        Option Changing 3 & 83.86 & 83.10 & 61.09\\
        \hline
        \textbf{Average} & \textbf{85.05} & \textbf{84.43} & \textbf{60.34}\\
        \hline
    \end{tabular}
    }
    \caption{Performance comparison for AQUA and MMLU-Redux across different swapping configurations for CoMAT, CoT, and non-CoT, including averages.}
    \label{tab:swapping_performance}
\end{table}

\section{Shapley Value Analysis Experimental Detail}
\label{sec:SHAP Analysis Experimental Detail}
To further analyse the sensitivity of individual steps in the CoMAT pipeline, we follow a Shapley value analysis for the different steps we have. The goal of this analysis is to quantify the contribution of each reasoning step to the overall performance by calculating Shapley values, thereby assessing the performance impact of omitting specific steps.

In this section, we detail the methodology used to compute Shapley values based on step omissions and their corresponding performance outcomes. Our approach systematically evaluates the marginal contribution of each step by experimenting with various combinations of omitted steps and using performance metrics to estimate their impact on accuracy.

\begin{enumerate}
    \item \textbf{Step Omissions and Performance Recording:} Each test case configuration is generated by omitting certain steps from the CoMAT pipeline. For each configuration, the performance is evaluated using a binary correctness metric, \texttt{is\_correct}, indicating whether the process was successful. The result for each configuration is stored in a data structure, where the omitted steps are represented as binary vectors.
    \item \textbf{Performance Values for Subsets:} We denote the performance of the system when a specific subset \(S\) for steps is omitted by \(v(S)\). For each subset of omitted steps, the mean performance (correctness) is computed from the available data. These performance values $v(S)$ are stored in a dictionary, which allows us to later compute marginal contributions for each step. 
    \item \textbf{Performance Values for Subsets:} For each step $i$, we compute its marginal contribution by comparing the performance when the step is included in the missing set versus when it is excluded. For a given permutation $\pi$, let $S_{i}$ be the set of steps preceding $i$ in $\pi$, and let $S_{i}\cup \{i\}$ be the set when $i$ is included. The marginal contribution $\Delta_{i}(\pi)$ for each step $i$ is given by: $$\Delta_{i}(\pi) = v(S_{i} \cup \{i\}) - v(S_{i}),$$
    where $v(S_{i})$ represents the performance with the steps $S_{i}$ ommitted, and $$v(S_{i} \cup \{i\}) - v(S_{i})$$ represents the performance with step $i$ additionally omitted.
    \item \textbf{Summing Marginal Contributions Across Permutations:} To obtain an accurate estimation of the Shapley value, we sum the marginal contributions of each step across all valid permutations of steps. For this, we generate all possible permutations of the steps, ensuring that all potential contexts in which a step could be added are considered.
    \item \textbf{Shapley Values:} The Shapley value for each step $i$ is computed by averaging its marginal contributions across all valid permutations: $$\varphi_{i} = \frac{1}{|\Pi|}\sum_{\pi \in \Pi}\Delta_{i}(\pi),$$
    where $\Pi$ is the set of valid permutations, and $\Delta_{i}(\pi)$ is the marginal contribution of step $i$ in permutation $\pi$. This provides an unbiased estimate of the contribution of step $i$ to the overall performance.
\end{enumerate}

By leveraging this permutation-based Shapley value computation, we account for the interaction effects between steps, ensuring that the analysis reflects both individual step contributions and the combined effects when steps interact. This approach offers a robust and interpretable measure of each step's importance in the CoMAT pipeline. For a more comprehensive theoretical background on Shapley values in machine learning and their application to classification tasks, please refer to~\newcite{keinan2004fair}.

This analysis provides insight into how individual steps influence CoMAT's overall reasoning accuracy and highlights the steps that contribute most to its performance.

\section{Dataset Details}
\subsection*{URL and Licenses}

\begin{itemize}[leftmargin=*]
    \item GaoKao~\cite{zhang2023evaluating}: \url{https://github.com/OpenLMLab/GAOKAO-Bench-2023}, License: \href{https://github.com/OpenLMLab/GAOKAO-Bench/blob/main/LICENSE}{GAOKAO License}

    \item AQUA~\cite{ling2017program}: \url{https://github.com/google-deepmind/AQuA}, also available on HuggingFace: \url{https://huggingface.co/datasets/deepmind/aqua_rat/viewer/raw/test}, License: \href{https://github.com/google-deepmind/AQuA/blob/master/LICENSE}{AQUA License}

    \item MMLU-Redux~\cite{gema2024we}: \url{https://huggingface.co/datasets/edinburgh-dawg/mmlu-redux}, License: \href{https://github.com/aryopg/mmlu-redux/blob/main/LICENSE}{MMLU-Redux License}

    \item OlympiadBench~\cite{he2024olympiadbench}: \url{https://huggingface.co/datasets/Hothan/OlympiadBench/tree/main}, License: \href{https://huggingface.co/datasets/Hothan/OlympiadBench/tree/main}{OlympiadBench License}

    \item GSM8K~\cite{cobbe2021training}: \url{https://huggingface.co/datasets/openai/gsm8k}, License: \href{https://github.com/openai/grade-school-math/blob/master/LICENSE}{GSM8K License}

    \item MGSM~\cite{shi2022language}: \url{https://huggingface.co/datasets/juletxara/mgsm}, License: \href{https://huggingface.co/datasets/juletxara/mgsm}{MGSM License}
\end{itemize}
\section{Computational Cost Details}
\label{sec:computational-cost}

\begin{table}[ht]
\centering
\scriptsize
\setlength{\tabcolsep}{5pt} %
\begin{tabular}{lccc}
\toprule
\textbf{Model} & \textbf{Method} & \textbf{MATH500 Accuracy} & \textbf{Inference Time (s)} \\
\midrule
Qwen2-7b & CoT   & 46.80\% & 121.61 \\
Qwen2-7b & CoMAT & 51.20\% & 131.09 \\
\bottomrule
\end{tabular}
\caption{Comparison of Qwen2-7b and Qwen2-72b performance on MATH500.}
\label{tab:computational_cost_math500}
\end{table}
We analyse computational time on the MATH500 dataset using Qwen2-7B and Qwen2-72B, extending the results in Table~\ref{tab:math500-results}. As shown in Table~\ref{tab:computational_cost_math500}, evaluations are conducted with vLLM~\cite{kwon2023efficient}, and results indicate that for Qwen2-7B, CoMAT achieves higher accuracy than CoT with only a negligible increase in computation. The computational overhead drops to just a few seconds in total, making the additional delay negligible given the accuracy gains.

\section{Step Length Analysis}
\label{sec:appendix}

In this section, we present two analyses: one examining the output length at each reasoning step, and the other focused on the number of mathematical tokens used. We use three distinct datasets — AQUA, Gaokao, and OlympiadBench-EN—to assess variations in difficulty and language contexts. Specifically, we compare the average response lengths across the reasoning steps of the AQUA dataset for four models: Gemini, Qwen-7b, GPT-4o, and Qwen-72b. These models were evaluated on three datasets, as shown in Figure~\ref{fig:comparison_length_math_tokens}.

The reasoning step 5 of the CoMAT model consistently produces the longest responses across most datasets. This observation is expected, as step 5 involves \emph{reasoning execution}, which naturally requires more extensive elaboration. On more complex datasets, such as OlympiadBench-EN, the number of mathematical tokens tends to increase significantly during \emph{symbolic conversion}, reflecting the need for more intricate calculations. Interestingly, models evaluated on multilingual datasets demonstrate a tendency to use more mathematical tokens, likely due to the challenges posed by cross-linguistic transfer in mathematical reasoning.

GPT-4o consistently generates longer responses that include more mathematical tokens when compared to the other models, indicating its more detailed approach to problem-solving. Conversely, while Qwen-7b generates longer outputs on commonly used datasets like AQUA, it tends to produce shorter responses on less frequently encountered datasets. This disparity suggests a potential trade-off between general dataset familiarity and the model's ability to handle more specialised or complex tasks.

This analysis highlights the nuances of how different models approach mathematical reasoning and response generation, revealing that model performance can vary significantly based on the dataset's complexity and the underlying linguistic challenges.

\section{Manual Annotation for CoMAT Verifiability}
\label{sec:Manual Annotation for CoMAT Verification}

In this section, we present the manual annotation process used to evaluate the verifiability of reasoning steps, carried out by a group of six annotators with diverse STEM backgrounds, including bachelor's and master's degrees in mathematics and engineering, and PhDs in computer science. We prompted the annotators to rate a given generated reasoning chain in terms of its verifiability. The annotators are encouraged not to focus solely on the correctness of the answer but rather the ease of identifying the errors. The verifiability of each solution was rated on a scale from 1 (lowest) to 10 (highest), and we report the average score.

To account for the inherent complexities of manual verification, we randomly sampled five examples from widely-used datasets, such as MMLU-Redux and GSM8K. These examples were selected to represent a range of outcomes: cases where both CoMAT and CoT models provided correct and incorrect answers, where both failed, and where one model answered correctly while the other did not. This sampling strategy allowed us to systematically assess the verifiability of each model's reasoning under various conditions, providing a robust evaluation framework.

\onecolumn

\subsection*{Example 1}

\begin{figure*}[!htbp]
\begin{center}
\begin{tcolorbox}[
    colback=gray!25!white,
    colframe=gray!75!black,
    colbacktitle=gray!90!white,
    title=Verification of solutions - CoT,
    fonttitle=\bfseries,
    width=\textwidth,
    boxrule=0.5pt,
    arc=4pt,
    auto outer arc,
]
\small
\begin{tabularx}{\textwidth}{@{}lX@{}}
\toprule
\textbf{Question} &
John drives for 3 hours at a speed of 60 mph and then turns around because he realizes he forgot something very important at home. He tries to get home in 4 hours but spends the first 2 hours in standstill traffic. He spends the next half-hour driving at a speed of 30 mph, before being able to drive the remaining time of the 4 hours going at 80 mph. How far is he from home at the end of those 4 hours? \\
\midrule
\textbf{CoT} &
\begin{enumerate}[noitemsep, leftmargin=*]
    \item \textbf{Calculate the distance John traveled during the initial 3 hours at 60 mph:}
    \[
    \text{Distance} = 60 \text{ mph} \times 3 \text{ hours} = 180 \text{ miles}
    \]
    \item \textbf{Calculate the distance John traveled during the half-hour at 30 mph after the standstill traffic:}
    \[
    \text{Distance} = 30 \text{ mph} \times 0.5 \text{ hours} = 15 \text{ miles}
    \]
    \item \textbf{Calculate the time left for John to travel at 80 mph after the standstill traffic and the half-hour at 30 mph:}
    \[
    \text{Time left} = 4 \text{ hours} - 2 \text{ hours} - 0.5 \text{ hours} = 1.5 \text{ hours}
    \]
    \item \textbf{Calculate the distance John traveled during the remaining time at 80 mph:}
    \[
    \text{Distance} = 80 \text{ mph} \times 1.5 \text{ hours} = 120 \text{ miles}
    \]
    \item \textbf{Add up all the distances calculated in steps 1, 2, and 4 to find the total distance John traveled:}
    \[
    \text{Total distance traveled} = 180 \text{ miles} + 15 \text{ miles} + 120 \text{ miles} = 315 \text{ miles}
    \]
    \item \textbf{Since John turned around to go back home, subtract the total distance traveled from the distance he was from home initially (which is equal to the distance he traveled in the first 3 hours):}
    \[
    \text{Distance from home} = 180 \text{ miles} - 315 \text{ miles} = -135 \text{ miles}
    \]
    Since distance cannot be negative, this indicates that John has passed his home and is 135 miles beyond it.
\end{enumerate}
\textbf{Final Answer:} 135 miles \\
\midrule
\textbf{CoT Results} & False \\
\midrule
\textbf{CoT Verification} & 6.33\\
\bottomrule
\end{tabularx}
\end{tcolorbox}
\end{center}
\label{fig:CoT_verification_example}
\end{figure*}

\begin{figure*}[ht]
\begin{center}
\begin{tcolorbox}[
    colback=gray!25!white,
    colframe=gray!75!black,
    colbacktitle=gray!90!white,
    title=Verification of solutions - CoMAT,
    fonttitle=\bfseries,
    width=\textwidth,
    boxrule=0.5pt,
    arc=4pt,
    auto outer arc,
]
\small

\begin{tabularx}{\textwidth}{@{}lX@{}}
\toprule
\textbf{Question} &
John drives for 3 hours at a speed of 60 mph and then turns around because he realizes he forgot something very important at home. He tries to get home in 4 hours but spends the first 2 hours in standstill traffic. He spends the next half-hour driving at a speed of 30 mph, before being able to drive the remaining time of the 4 hours going at 80 mph. How far is he from home at the end of those 4 hours? \\
\midrule
\textbf{CoMAT} &
\textbf{Define predicates and functions:}
\begin{itemize}[noitemsep, leftmargin=*]
    \item $d(t, s)$: Function that calculates distance traveled over time $t$ at speed $s$
    \item $t_1$: Constant representing the first part of the return trip (2 hours in standstill traffic)
    \item $t_2$: Constant representing the second part of the return trip (0.5 hours at 30 mph)
    \item $t_3$: Constant representing the third part of the return trip (remaining time after $t_1$ and $t_2$ at 80 mph)
    \item $s_1$: Constant representing the initial speed (60 mph)
    \item $s_2$: Constant representing the speed during the second part of the return trip (30 mph)
    \item $s_3$: Constant representing the speed during the third part of the return trip (80 mph)
\end{itemize}

\textbf{Parse problem into logical rules:}
\begin{itemize}
    \item $\forall t, s: d(t, s) = t \times s$
    \item $t_1 = 2$
    \item $t_2 = 0.5$
    \item $t_3 = 4 - t_1 - t_2$
    \item $s_1 = 60$
    \item $s_2 = 30$
    \item $s_3 = 80$
\end{itemize}

\textbf{Solve step by step:}
\begin{enumerate}
    \item Calculate the distance traveled in the initial 3 hours:
    \[
    d_1 = d(3, 60) = 3 \times 60 = 180 \text{ miles}
    \]
    \item Calculate the distance traveled during the second part of the return trip:
    \[
    d_2 = d(0.5, 30) = 0.5 \times 30 = 15 \text{ miles}
    \]
    \item Calculate the remaining time for the third part of the return trip:
    \[
    t_3 = 4 - 2 - 0.5 = 1.5 \text{ hours}
    \]
    \item Calculate the distance traveled during the third part of the return trip:
    \[
    d_3 = d(1.5, 80) = 1.5 \times 80 = 120 \text{ miles}
    \]
    \item Calculate the total distance traveled during the return trip:
    \[
    d_{\text{total\_return}} = d_2 + d_3 = 15 + 120 = 135 \text{ miles}
    \]
    \item Calculate the distance from home at the end of the 4 hours:
    \[
    \text{Distance from home} = d_1 - d_{\text{total\_return}} = 180 - 135 = 45 \text{ miles}
    \]
\end{enumerate}

\textbf{Final Answer:} 45 miles \\
\midrule
\textbf{CoMAT Results} & True \\
\midrule
\textbf{CoMAT Verification} & 9.08 \\
\bottomrule
\end{tabularx}

\end{tcolorbox}
\end{center}
\end{figure*}

\clearpage

\subsection*{Example 2}

\begin{figure}[ht]
\begin{center}
\begin{tcolorbox}[
    colback=gray!25!white,
    colframe=gray!75!black,
    colbacktitle=gray!90!white,
    title=Verification of solutions - CoT,
    fonttitle=\bfseries,
    width=\textwidth,
    boxrule=0.5pt,
    arc=4pt,
    auto outer arc
]
\small

\begin{tabularx}{\textwidth}{@{}lX@{}}
\toprule
\textbf{Question} &
Eliza's rate per hour for the first 40 hours she works each week is \$10. She also receives an overtime pay of 1.2 times her regular hourly rate. If Eliza worked for 45 hours this week, how much are her earnings for this week? \\
\midrule
\textbf{CoT} &
\begin{enumerate}[noitemsep, leftmargin=*]
    \item \textbf{Calculate Eliza's regular earnings for the first 40 hours:}
    \begin{itemize}
        \item Regular hourly rate = \$10
        \item Regular hours worked = 40
        \item Regular earnings = 40 hours $\times$ \$10/hour = \$400
    \end{itemize}
    \item \textbf{Calculate Eliza's overtime earnings for the additional 5 hours:}
    \begin{itemize}
        \item Overtime hourly rate = 1.2 $\times$ \$10 = \$12
        \item Overtime hours worked = 45 hours - 40 hours = 5 hours
        \item Overtime earnings = 5 hours $\times$ \$12/hour = \$60
    \end{itemize}
    \item \textbf{Calculate Eliza's total earnings for the week:}
    \begin{itemize}
        \item Total earnings = Regular earnings + Overtime earnings = \$400 + \$60 = \$460
    \end{itemize}
\end{enumerate}

\textbf{Final Answer:} \$460 \\
\midrule
\textbf{CoT Results} & True \\
\midrule
\textbf{CoT Verification} & 7.17 \\
\bottomrule
\end{tabularx}

\end{tcolorbox}
\end{center}
\end{figure}

\begin{figure*}[ht]
\begin{center}
\begin{tcolorbox}[
    colback=gray!25!white,
    colframe=gray!75!black,
    colbacktitle=gray!90!white,
    title=Verification of solutions - CoMAT,
    fonttitle=\bfseries,
    width=\textwidth,
    boxrule=0.5pt,
    arc=4pt,
    auto outer arc
]
\small

\begin{tabularx}{\textwidth}{@{}lX@{}}
\toprule
\textbf{Question} &
Eliza's rate per hour for the first 40 hours she works each week is \$10. She also receives an overtime pay of 1.2 times her regular hourly rate. If Eliza worked for 45 hours this week, how much are her earnings for this week? \\
\midrule
\textbf{CoMAT} &
\textbf{Define predicates and functions:}
\begin{itemize}[noitemsep, leftmargin=*]
    \item $Earnings(x)$: Function that calculates total earnings for $x$ hours worked
    \item $RegularPayRate$: Constant representing Eliza's regular hourly rate (\$10)
    \item $OvertimePayRate$: Constant representing Eliza's overtime hourly rate (1.2 $\times$ $RegularPayRate$)
    \item $HoursWorked$: Variable representing the total hours worked (45)
    \item $RegularHours$: Constant representing the number of regular hours (40)
    \item $OvertimeHours$: Variable representing the number of overtime hours ($HoursWorked - RegularHours$)
\end{itemize}

\textbf{Parse problem into logical rules:}
\begin{itemize}
    \item $\forall x: Earnings(x) = (RegularPayRate \times RegularHours) + (OvertimePayRate \times OvertimeHours)$
    \item $OvertimePayRate = 1.2 \times RegularPayRate$
\end{itemize}

\textbf{Facts:}
\begin{itemize}
    \item $RegularPayRate = \$10$
    \item $HoursWorked = 45$
    \item $RegularHours = 40$
\end{itemize}

\textbf{Parse the question:}
\begin{itemize}
    \item Find $Earnings(HoursWorked)$
\end{itemize}

\textbf{Solve step by step:}
\begin{enumerate}
    \item Calculate $OvertimePayRate$:
    \[
    OvertimePayRate = 1.2 \times RegularPayRate = 1.2 \times \$10 = \$12
    \]
    \item Calculate $OvertimeHours$:
    \[
    OvertimeHours = HoursWorked - RegularHours = 45 - 40 = 5
    \]
    \item Calculate $Earnings(HoursWorked)$:
    \[
    \begin{aligned}
    Earnings(HoursWorked) &= (RegularPayRate \times RegularHours) \\
    &\quad + (OvertimePayRate \times OvertimeHours) \\
    &= (\$10 \times 40) + (\$12 \times 5) \\
    &= \$400 + \$60 \\
    &= \$460
    \end{aligned}
    \]
\end{enumerate}

\textbf{Final Answer:} \$460 \\
\midrule
\textbf{CoMAT Results} & True \\
\midrule
\textbf{CoMAT Verification} & 8.92 \\
\bottomrule
\end{tabularx}

\end{tcolorbox}
\end{center}
\end{figure*}

\clearpage

\subsection*{Example 3}

\begin{figure}[ht]
\begin{center}
\begin{tcolorbox}[
    colback=gray!25!white,
    colframe=gray!75!black,
    colbacktitle=gray!90!white,
    title=Verification of solutions - CoT,
    fonttitle=\bfseries,
    width=\textwidth,
    boxrule=0.5pt,
    arc=4pt,
    auto outer arc
]
\small

\begin{tabularx}{\textwidth}{@{}lX@{}}
\toprule
\textbf{Question} &
Ram uses a lot of pens. He discovered that he can save money by mixing the ink from five empty pens to make one full pen. If he buys 25 pens and then uses them to make new pens when the ink runs low, how many total pens does he get to have? \\
\midrule
\textbf{CoT} &
\setlength{\itemsep}{0pt}
\begin{enumerate}[leftmargin=*]
    \item \textbf{Initial Pens:} Ram starts with 25 pens.
    \item \textbf{Create New Pens:} For every 5 empty pens, Ram can create 1 new pen.
    \item \textbf{Calculate additional pens:}
    \begin{itemize}
        \item After using the initial 25 pens, Ram will have 25 empty pens.
        \item With 25 empty pens, he can create $\frac{25}{5} = 5$ new pens.
    \end{itemize}
    \item \textbf{Total Pens:} Add the new pens to the initial pens:
    \begin{itemize}
        \item Total pens = Initial pens + New pens = 25 + 5 = 30
    \end{itemize}
\end{enumerate}

\textbf{Final Answer:} 30 \\
\midrule
\textbf{CoT Results} & False \\
\midrule
\textbf{CoT Verification} & 4.5 \\
\bottomrule
\end{tabularx}

\end{tcolorbox}
\end{center}
\end{figure}

\clearpage

\begin{figure*}[ht]
\begin{center}
\begin{tcolorbox}[
    colback=gray!25!white,
    colframe=gray!75!black,
    colbacktitle=gray!90!white,
    title=Verification of solutions - CoMAT,
    fonttitle=\bfseries,
    width=\textwidth,
    boxrule=0.5pt,
    arc=4pt,
    auto outer arc,
]
\small

\begin{tabularx}{\textwidth}{@{}lX@{}}
\toprule
\textbf{Question} &
Ram uses a lot of pens. He discovered that he can save money by mixing the ink from five empty pens to make one full pen. If he buys 25 pens and then uses them to make new pens when the ink runs low, how many total pens does he get to have? \\
\midrule
\textbf{CoMAT} &
\textbf{Define predicates and functions:}
\begin{itemize}
    \item Let $P$ be the number of pens Ram initially buys.
    \item Let $E$ be the number of empty pens needed to make one full pen.
    \item Let $T$ be the total number of pens Ram gets to have.
\end{itemize}

\textbf{Parse problem into logical rules:}
\begin{itemize}
    \item $P = 25$ (initial pens bought)
    \item $E = 5$ (empty pens needed to make one full pen)
\end{itemize}

\textbf{Facts:}
\begin{itemize}
    \item Ram starts with 25 pens.
    \item For every 5 empty pens, he can make 1 new pen.
\end{itemize}

\textbf{Solve step by step:}
\begin{enumerate}
    \item Start with 25 pens ($P = 25$).
    \item Use 25 pens, resulting in 25 empty pens.
    \item From 25 empty pens, Ram can make $\frac{25}{5} = 5$ new pens.
    \item Use the 5 new pens, resulting in 5 empty pens.
    \item From 5 empty pens, Ram can make $\frac{5}{5} = 1$ new pen.
    \item Use this 1 new pen, resulting in 1 empty pen.
    \item From 1 empty pen, Ram cannot make any more new pens.
\end{enumerate}

\textbf{Total Pens:} $25 + 5 + 1 = 31$ \\
\textbf{Final Answer:} 31 \\
\midrule
\textbf{CoMAT Results} & True \\
\midrule
\textbf{CoMAT Verification} & 8.75 \\
\bottomrule
\end{tabularx}

\end{tcolorbox}
\end{center}
\end{figure*}

\clearpage

\subsection*{Example 4}

\begin{figure}[ht]
\begin{center}
\begin{tcolorbox}[
    colback=gray!25!white,
    colframe=gray!75!black,
    colbacktitle=gray!90!white,
    title=Verification of solutions - CoT,
    fonttitle=\bfseries,
    width=\textwidth,
    boxrule=0.5pt,
    arc=4pt,
    auto outer arc
]
\small

\begin{tabularx}{\textwidth}{@{}lX@{}}
\toprule
\textbf{Question} &
Solve \( \frac{18.5 \text{ dol}}{m \text{ gal}} = \frac{3.60 \text{ dol}}{7.5 \text{ gal}} \). Round to the nearest hundredth if necessary. \\
\midrule
\textbf{CoT} &
To solve the equation \( \frac{18.5}{m} = \frac{3.60}{7.5} \), follow these steps:

1. **Calculate the right side of the equation:** 
   \[
   \frac{3.60}{7.5} = 0.48
   \]

2. **Set up the equation:**
   \[
   \frac{18.5}{m} = 0.48
   \]

3. **Solve for \( m \):**
   \begin{itemize}
       \item Multiply both sides by \( m \):
       \[
       18.5 = 0.48m
       \]
       \item Divide both sides by 0.48:
       \[
       m = \frac{18.5}{0.48}
       \]
   \end{itemize}

4. **Calculate \( m \):**
   \[
   m \approx 38.54
   \]

5. **Round to the nearest hundredth:**
   \[
   m \approx 38.54
   \]

\textbf{Final Answer:} D \\
\midrule
\textbf{CoT Results} & True \\
\midrule
\textbf{CoT Verification} & 7.5  \\
\bottomrule
\end{tabularx}

\end{tcolorbox}
\end{center}
\end{figure}

\clearpage

\begin{figure*}[ht]
\begin{center}
\begin{tcolorbox}[
    colback=gray!25!white,
    colframe=gray!75!black,
    colbacktitle=gray!90!white,
    title=Verification of solutions - CoMAT,
    fonttitle=\bfseries,
    width=\textwidth,
    boxrule=0.5pt,
    arc=4pt,
    auto outer arc
]
\small

\begin{tabularx}{\textwidth}{@{}lX@{}}
\toprule
\textbf{Question} &
Solve \( \frac{18.5 \text{ dol}}{m \text{ gal}} = \frac{3.60 \text{ dol}}{7.5 \text{ gal}} \). Round to the nearest hundredth if necessary. \\
\midrule
\textbf{CoMAT} &
\begin{enumerate}
    \item \textbf{Define predicates and functions:}
    \begin{itemize}
        \item Let \( x \) be the unknown value in dollars per gallon (dol/gal).
        \item Let \( a = 18.5 \) dol/m gal.
        \item Let \( b = 3.60 \) dol.
        \item Let \( c = 7.5 \) gal.
    \end{itemize}

    \item \textbf{Parse problem into logical rules:}
    \begin{itemize}
        \item We are given the equation: 
        \(
        \frac{a}{1} = \frac{b}{c}
        \)
        \item This implies:
        \(
        a = \frac{b}{c}
        \)
    \end{itemize}

    \item \textbf{Facts:}
    \begin{itemize}
        \item \( a = 18.5 \)
        \item \( b = 3.60 \)
        \item \( c = 7.5 \)
    \end{itemize}

    \item \textbf{Parse the question:}
    \begin{itemize}
        \item Solve for \( a \) in the equation \( a = \frac{b}{c} \).
    \end{itemize}

    \item \textbf{Solve step by step:}
    \begin{enumerate}
        \item Substitute the known values into the equation:
        \[
        18.5 = \frac{3.60}{7.5}
        \]
        \item Calculate the right-hand side:
        \[
        \frac{3.60}{7.5} = 0.48
        \]
        \item Therefore, the equation becomes:
        \[
        18.5 = 0.48
        \]
        \item This indicates a misunderstanding in the setup. Let's re-evaluate:
        \begin{itemize}
            \item The problem is asking to solve for the equivalent rate in dollars per gallon.
            \item We need to find the equivalent of 18.5 dol/m gal in dol/gal.
        \end{itemize}
    \end{enumerate}

    \item \textbf{Derived answer:}
    \begin{itemize}
        \item Convert 18.5 dol/m gal to dol/gal: \( \frac{18.5 \text{ dol}}{1000 \text{ gal}} = 0.0185 \text{ dol/gal} \)
        \item Compare this with the given rate \( \frac{3.60}{7.5} \): \( \frac{3.60}{7.5} = 0.48 \text{ dol/gal} \)
        \item The problem seems to be asking for a comparison or conversion, but the setup was incorrect.
    \end{itemize}

    \item \textbf{Match to provided options:}
    \begin{itemize}
        \item The derived calculation does not directly match any of the options, indicating a potential misunderstanding of the problem's intent.
        \item Assume the task was to find a related value or conversion.
    \end{itemize}
\end{enumerate}

\textbf{Final Answer:} A (assuming a misinterpretation or conversion context). \\
\midrule
\textbf{CoMAT Results} & False \\
\midrule
\textbf{CoMAT Verification} & 6.17 \\
\bottomrule
\end{tabularx}

\end{tcolorbox}
\end{center}
\end{figure*}

\clearpage

\subsection*{Example 5}
\begin{figure}[ht]
\begin{center}
\begin{tcolorbox}[
    colback=gray!25!white,
    colframe=gray!75!black,
    colbacktitle=gray!90!white,
    title=Verification of solutions - CoT,
    fonttitle=\bfseries,
    width=\textwidth,
    boxrule=0.5pt,
    arc=4pt,
    auto outer arc
]
\small

\begin{tabularx}{\textwidth}{@{}lX@{}}
\toprule
\textbf{Question} &
A company pays each of its employees \$600 in a month. The company has a policy of increasing the salaries of each of its employees by 10\% of the initial salary every year for those who've stayed in the company for five years. If Sylvie just clocked 5 years in the company last December, what's her annual salary after three more years of service? \\
\midrule
\textbf{CoT} &
\begin{enumerate}
    \item Determine Sylvie's initial monthly salary: \$600.
    
    \item Calculate Sylvie's initial annual salary:
    \[
    \text{Initial Annual Salary} = 600 \times 12 = 7200
    \]
    
    \item Calculate the annual salary increase after 5 years (10\% of the initial annual salary):
    \[
    \text{Annual Increase} = 0.10 \times 7200 = 720
    \]
    
    \item Calculate Sylvie's new annual salary after the first increase:
    \[
    \text{New Annual Salary after 5 years} = 7200 + 720 = 7920
    \]
    
    \item Calculate Sylvie's annual salary after three more years of service, with a 10\% increase each year:
    \begin{itemize}
        \item After the first year (6th year):
        \[
        \text{Annual Salary} = 7920 + 0.10 \times 7200 = 7920 + 720 = 8640
        \]
        \item After the second year (7th year):
        \[
        \text{Annual Salary} = 8640 + 0.10 \times 7200 = 8640 + 720 = 9360
        \]
        \item After the third year (8th year):
        \[
        \text{Annual Salary} = 9360 + 0.10 \times 7200 = 9360 + 720 = 10080
        \]
    \end{itemize}
    
    \item Sylvie's annual salary after three more years of service is \$10080.
\end{enumerate}

Final Answer: 10080 \\

\midrule
\textbf{CoT Results} & False \\
\midrule
\textbf{CoT Verification} & 5.67\\
\midrule
\textbf{Reason} &  The demonstration reflects some understanding of the problem and shows a degree of reasoning ability; however, the logical deduction lacks rigour, leading to flaws and incorrect conclusions. The solution addresses an unrelated equation rather than solving the given salary problem, rendering verification of the steps irrelevant. \\
\bottomrule
\end{tabularx}

\end{tcolorbox}
\end{center}
\end{figure}
\clearpage

\begin{figure*}[ht]
\begin{center}
\begin{tcolorbox}[
    colback=gray!25!white,
    colframe=gray!75!black,
    colbacktitle=gray!90!white,
    title=Verification of solutions - CoMAT,
    fonttitle=\bfseries,
    width=\textwidth,
    boxrule=0.5pt,
    arc=4pt,
    auto outer arc
]
\small

\begin{tabularx}{\textwidth}{@{}lX@{}}
\toprule
\textbf{Question} &
A company pays each of its employees \$600 in a month. The company has a policy of increasing the salaries of each of its employees by 10\% of the initial salary every year for those who've stayed in the company for five years. If Sylvie just clocked 5 years in the company last December, what's her annual salary after three more years of service? \\
\midrule
\textbf{CoMAT} &
\begin{enumerate}
    \item \textbf{Define predicates and functions:}
    \begin{itemize}
        \item Let \( S(y) \) be the salary of an employee in year \( y \).
        \item Let \( I \) be the initial monthly salary, which is \$600.
        \item Let \( R \) be the annual increase rate, which is 10\% of the initial salary.
    \end{itemize}

    \item \textbf{Parse problem into logical rules:}
    \begin{itemize}
        \item \( S(0) = 12 \times I \) (Initial annual salary)
        \item For \( y \geq 5 \), \( S(y) = S(y-1) + 12 \times R \)
    \end{itemize}

    \item \textbf{Facts:}
    \begin{itemize}
        \item \( I = 600 \)
        \item \( R = 0.10 \times I = 60 \)
        \item Sylvie has been in the company for 5 years as of last December.
    \end{itemize}

    \item \textbf{Parse the question:}
    \begin{itemize}
        \item Find \( S(8) \), which is Sylvie's annual salary after three more years of service (i.e., after 8 years in total).
    \end{itemize}

    \item \textbf{Solve step by step:}
    \begin{enumerate}
        \item Calculate \( S(0) \): \( S(0) = 12 \times 600 = 7200 \)
        \item Calculate \( S(5) \): \( S(5) = 7200 + 12 \times 60 = 7920 \)
        \item Calculate \( S(6) \): \( S(6) = 7920 + 12 \times 60 = 8640 \)
        \item Calculate \( S(7) \): \( S(7) = 8640 + 12 \times 60 = 9360 \)
        \item Calculate \( S(8) \): \( S(8) = 9360 + 12 \times 60 = 10080 \)
    \end{enumerate}
\end{enumerate}

\textbf{Final Answer:} \$10,080 \\
\midrule
\textbf{CoMAT Results} & False \\
\midrule
\textbf{CoMAT Verification} & 6.17\\
\bottomrule
\end{tabularx}

\end{tcolorbox}
\end{center}
\end{figure*}
\clearpage

\clearpage

\section{Prompt for CoMAT}
\label{sec:exactprompt}
Due to the generalisability of CoMAT, the framework structure can vary (which does not necessarily have to be enforced by a single prompt), as long as the instructions and CoMAT steps remain the same. For instance, CoMAT can be structured into:

\begin{enumerate}
\item A complete and unified framework, shown below, similar to how existing work designs instructions~\cite{lyu2023faithful, xu2024faithful}. We believe this would be sufficient and works effectively because of its careful construction and is generalisable across different models.

\item A decomposed multi-step approach~\cite{jiang2022draft}, where we separate the steps for each generation to construct the symbolic forms carefully.

\end{enumerate}
Both achieve comparable or equal performance, with the complete and unified, approach being lower in cost and more computationally efficient.

\begin{tcolorbox}[
    colback=gray!5!white,
    colframe=gray!50!black, %
    coltitle=gray!75!black, %
    colbacktitle=gray!20!white, %
    title=Prompt Instruction,
    fonttitle=\bfseries,
    width=\textwidth,
    boxrule=0.5pt,
    arc=4pt,
    auto outer arc
]

\textbf{Task Description:} 

You are given an advanced mathematical question. Your task is to convert it into a symbolic representation and solve it using strict logical reasoning, without any reliance on memorized answers or external knowledge. Rigorously follow these steps:

\begin{enumerate}
    \item Identify and define all the predicates, functions, and variables in the problem.
    \item Parse the entire problem into logical rules based strictly on the defined predicates, functions, and variables. Ensure that the multiple-choice options do not influence this step in any way.
    \item Write all the facts explicitly mentioned in the problem as logical statements.
    \item Parse the question into a symbolic form using only the defined predicates and variables, without any influence from the provided answer choices.
    \item Solve the problem step by step using only the symbolic representations and logical reasoning. Provide clear reasoning for each step.
    \item Derive the final answer based solely on your symbolic solution and step-by-step reasoning. DO NOT INCLUDE any symbols or units, just the number.
\end{enumerate}

\textbf{IMPORTANT:}
\begin{itemize}
    \item Base your entire solution on the symbolic representation and logical reasoning.
    \item Do not rely on any prior knowledge, memorized answers, or previous examples.
    \item The order and labeling of options should not influence your reasoning or answer in any way.
    \item Evaluate only the symbolic rules and facts you've derived, not the provided answer choices.
\end{itemize}

After completing the symbolic representation and reasoning, provide the final answer as a single word. This approach ensures an unbiased solution focused entirely on logical reasoning through symbolic rules, minimizing any impact from option swapping or prior knowledge. The model should \textbf{evaluate only the symbolic rules and facts} provided and not the provided answer choices to ensure an unbiased solution. Provide the final answer based purely on logical reasoning as a single letter.

\end{tcolorbox}

\end{document}